\documentclass{jdmdh}
\usepackage{array}
\usepackage{pgfplots}
\usepackage{textcomp}
\usepackage{amsmath}
\usepackage{booktabs}
\pgfplotsset{compat=1.18}
\usepackage{times}
\usepackage{latexsym}
\usepackage[T1]{fontenc}
\usepackage[utf8]{inputenc}

\usepackage{microtype}

\usepackage{inconsolata}
\usepackage{footnotehyper}

\titlespacing*{\section}
{0pt}{2.5ex plus 1ex minus .2ex}{1.3ex plus .2ex}
\titlespacing*{\subsection}
{0pt}{1.0ex plus 1ex minus .2ex}{1.3ex plus .2ex}
\titlespacing*{\subsubsection}
{0pt}{1.0ex plus 1ex minus .2ex}{1.3ex plus .2ex}

\title{Efficient Toxicity Detection in Gaming Chats: A Comparative Study of Embeddings, Fine-Tuned Transformers and LLMs}
\author{
  Yehor Tereshchenko and Mika Hämäläinen \\
  Metropolia University of Applied Sciences \\
  Helsinki, Finland \\
  \texttt{firstname.lastname@metropolia.fi} \\
}

\begin{document}

\maketitle

\abstract{This paper presents a comprehensive comparative analysis of Natural Language Processing (NLP) methods for automated toxicity detection in online gaming chats. Traditional machine learning models with embeddings, large language models (LLMs) with zero-shot and few-shot prompting, fine-tuned transformer models, and retrieval-augmented generation (RAG) approaches are evaluated. The evaluation framework assesses three critical dimensions: classification accuracy, processing speed, and computational costs. A hybrid moderation system architecture is proposed that optimizes human moderator workload through automated detection and incorporates continuous learning mechanisms. The experimental results demonstrate significant performance variations across methods, with fine-tuned DistilBERT achieving optimal accuracy-cost trade-offs. The findings provide empirical evidence for deploying cost-effective, efficient content moderation systems in dynamic online gaming environments.
}


\section{Introduction}
\label{sec:introduction}
Online gaming has evolved from isolated entertainment into a substantial social ecosystem where millions
of users engage in real-time communication.  
While this transformation facilitates engagement and community formation, it simultaneously exposes participants to pervasive
toxic language and abusive behavior \citep{Ghosh2021Analyzing}. Conventional manual moderation systems
struggle to maintain pace with the velocity, volume, and linguistic complexity of modern gaming communications,
frequently allowing harmful content to persist sufficiently long to degrade user experience and discourage participation from
marginalized groups \citep{Xiao2023Addressing}.

This research investigates how recent advances in Natural Language Processing (NLP)— from lightweight embedding-based classifiers to modern Large Language Models (LLMs) and
Retrieval-Augmented Generation (RAG)—can be systematically integrated into cost-effective, high-throughput, and
ethically responsible moderation workflows.

\subsection{Context and Motivation}
\label{subsec:context}
The global gaming market is projected to exceed \(\$250\) billion in revenue by 2025 , with chat-enabled
multiplayer titles constituting a significant proportion of user engagement \citep{newzoo2024global}.
These dynamic communities inevitably reflect broader societal challenges: hate speech, harassment,
profanity, threats, and discriminatory language proliferate, frequently exacerbated by in-game anonymity and
rapid linguistic evolution  \citep{Banks2010Regulating}.
Unaddressed, such content undermines user trust, discourages minority participation, and may violate
regional regulatory frameworks (\citealt{Blackwell2019Harassment}).  

Although human moderators represent the optimal standard for nuanced judgment, exclusive reliance on human moderation incurs
substantial operational costs, introduces significant response delays, and contributes to cognitive overload and
professional burnout (\citealt{MacAvaney2019Hate}, \citealt{Kovacs2021Challenges}). Automated assistance is therefore essential—not to replace human expertise, but to filter routine cases
so that human experts can focus on borderline judgments and policy refinement \citep{Kumar2024Watch}. 

\subsection{Problem Statement}
\label{subsec:problem_statement}
This research addresses the following question: \emph{"Which combination of contemporary NLP techniques delivers the best balance of accuracy,
speed, and resource efficiency for automated ethical validation of gaming chat messages?"}
Specifically, the following approaches are compared: (i) traditional machine-learning classifiers trained on static or contextual
embeddings, (ii) zero-/few-shot prompting of powerful foundation models, (iii) task-specific fine-tuning
of open-source LLMs, and (iv) LLMs enhanced with a RAG pipeline that supplies domain knowledge such as
previously moderated utterances.  The goal is to quantify performance trade-offs and to identify a
deployable hybrid design that minimises manual workload without compromising fairness or precision.

%\subsection{Paper Structure}
%\label{subsec:structure}
%Section \ref{sec:related_work} reviews prior work on hate-speech detection, text embeddings, and LLM-based moderation. Section \ref{sec:methodology} describes data collection, model configurations, and evaluation protocols. Section \ref{sec:experiments_results} presents the experimental results, which are analyzed in Section \ref{sec:discussion}. Section \ref{sec:conclusion} concludes with future research directions. The Limitations and Ethics Statement sections discuss broader ethical implications and study constraints.

\section{Related Work}
\label{sec:related_work}
This section provides a comprehensive overview of existing research and methodologies relevant to our investigation.
\subsection{Methods for Hate Speech and Undesirable Content Detection}
\label{subsec:hate_speech_detection}

Early automated moderation systems largely relied on rule-based filters, in which incoming chat messages were scanned against manually constructed blacklists of banned words or regular-expression patterns. Such approaches are straightforward to implement and incur minimal runtime cost, but suffer from severe brittleness. They cannot recognize context, fail to catch obfuscated or intentionally misspelled slurs (e.g., "h*te" for "hate"), and are wholly blind to sarcasm, irony, or evolving in-group slang—resulting in both high false-positive and high false-negative rates in dynamic gaming conversations (see \citealt{Govindankutty2024Design}, \citealt{MacAvaney2019Hate}).

To improve robustness, subsequent work introduced conventional machine-learning classifiers trained on engineered text features. Typical pipelines represent each message as a high-dimensional vector of TF-IDF–weighted unigrams, bigrams, or character n-grams, sometimes augmented with sentiment lexicon scores or part-of-speech tags \citep{Zhou2022Research}. These vectors are then fed into classifiers such as Support Vector Machines (SVMs), Naive Bayes, or Random Forests. (about vectors see in \citealt{Wibowo2024Performance}) On benchmark hate-speech datasets, such methods achieved high accuracy, outperforming purely lexical filters (\citealt{Das2023Performance}). However, they still struggle when the vocabulary shifts or when subtle syntactic cues signal malicious intent, since their static feature sets cannot generalize to unseen or cleverly disguised hateful expressions. \citep{Suwarno2021Comparative}

Over time, researchers enriched these pipelines with hand-crafted syntactic and semantic patterns (e.g., dependency-parse templates for insults or threats) and built ensemble systems that combine multiple feature types and classifiers to reduce individual weaknesses (for example see \citealt{Galitsky2016Improving}). While these enhancements yielded incremental gains, the reliance on manually maintained feature extractors and lexicons limited scalability: each new slang term or obfuscation tactic required fresh engineering effort \cite{Sun2022Semantically}.  This realization paved the way for embedding-based and neural architectures—methods capable of automatically learning dense, low-dimensional representations that can capture deeper semantic and contextual nuances without exhaustive manual feature design (see Section~\ref{subsec:text_embeddings}).
\subsection{The Role of Text Embeddings in NLP}
\label{subsec:text_embeddings}

Word embeddings map tokens (words or subwords) into dense, low-dimensional vectors such that semantically similar items lie close together in vector space \cite{Worth2023Word}. Early static methods—Word2Vec (for details see \citealt{Goldberg2014word2vec}), GloVe (for details and comparison Glove with Word2Vec see \citealt{Shi2014Linking}), and FastText (\citealt{Umer2022Impact})—train on large unlabeled corpora to capture global co-occurrence statistics or local context windows.  Such embeddings proved effective in many downstream tasks, including toxic speach detection (see \citealt{Malik2021Toxic}), but assign each word a single vector regardless of its usage, making it difficult to distinguish polysemy or context‐specific meaning (see \citealt{Arora2018Linear}).

Contextual embeddings address this by producing token representations that vary with surrounding text.  Models like BERT \cite{Devlin2018BERT}, RoBERTa \cite{Liu2019RoBERTa} , and XLNet \cite{Yang2019XLNet} pre-train deep bidirectional or autoregressive transformers on masked‐language or permutation objectives, then extract hidden‐state vectors as context‐aware word embeddings.  Sentence‐level variants (e.g., Sentence-BERT; \citealt{Reimers2019Sentence}) further refine these into fixed‐length vectors optimized for semantic similarity tasks.  In dynamic gaming chats—where slang evolves rapidly and sarcasm abounds—contextual embeddings significantly outperform static ones in capturing nuance. \citep{Naseem2020Towards}

Beyond text, user embeddings have also been explored.  \citet{hamalainen2024twitch} apply an LLM to generate "user embeddings" for Twitch chatters, clustering them into categories like supportive viewers and emoji-heavy spammers; this demonstrates how embedding techniques can model not just content, but speaker behavior in real time.  Likewise, \citet{poyhonen2022persuasion} extract multilingual RPG dialogue to train BERT on persuasion detection, showing that domain-specific embeddings can be quickly adapted to niche tasks such as identifying persuasive intent in game text.  

\subsection{Application of Large Language Models (LLMs) in Moderation}
\label{subsec:llm_moderation}

Large pre-trained LLMs (e.g., GPT-3.5\footnote{\label{fn:gpt35}\url{https://openai.com/blog/gpt-3-5-turbo-fine-tuning-and-api-updates}}/GPT-4\footnote{\label{fn:gpt4}\url{https://openai.com/research/gpt-4}}) have shown remarkable zero- and few-shot capabilities for classification by prompt engineering.  In zero-shot classification, a prompt such as "Decide whether the following message contains hate speech: '…'" elicits a direct yes/no answer, often with competitive accuracy compared to fine-tuned baselines.  Few-shot prompting further improves performance by providing a handful of labeled examples in the prompt to guide the model's reasoning. \citep{Gandhi2024Hate}

Fine-tuning LLMs on in-domain chat datasets adapts them to gaming-specific slang and abusive patterns.  For example, \citet{alnajjar2022emotion} fine-tuned DialoGPT\footnote{\label{fn:dialogpt}\url{https://huggingface.co/microsoft/DialoGPT-medium}} to generate emotion-conditioned dialog, demonstrating that relatively small, domain‐focused corpora can significantly steer a large model's output style .  Similarly, fine-tuned LLMs achieve higher precision and lower false-positive rates in moderation tasks, at the cost of additional training compute and maintenance \citep{2023Adapting}.

Retrieval-Augmented Generation (RAG) systems combine a dense retrieval component with an LLM.  Given an input message, the retriever fetches similar past utterances or policy documents from a vector store; the LLM then conditions on both the query and retrieved context to produce a classification or moderation verdict.  This approach reduces model "hallucinations" and allows the system to leverage fresh or proprietary moderation logs without retraining the LLM from scratch.  RAG is particularly promising for handling rapidly evolving game-specific memes or coded insults by consulting an external knowledge base of recent examples. \citep{Chen2024Class}

\subsection{Existing Automated Content Moderation Systems}
\label{subsec:existing_systems}

Major platforms and game publishers employ multistage pipelines the blend rule-based filters, ML classifiers, and human review.  For instance, Twitch's AutoMod uses keyword blacklists and heuristics to flag messages for volunteer moderators (see \citealt{Cai2019Categorizing}). Commercial solutions such as Google’s Perspective API provide commercially available toxicity scoring, though they often struggle to adapt to the fast paced, jargon filled language of gaming communities \cite{Hosseini2017Deceiving}.

Research prototypes, such as the system described by \citet{hamalainen2024twitch}, demonstrate the viability of embedding-based clustering for user behavior analysis in Twitch streams.  Yet even advanced systems report high false-positive rates when encountering novel slang or code words, and high false-negative rates when malicious users obfuscate content \cite{Fkih2023Machine}.  Scalability and latency constraints also limit the practicality of large LLMs in real-time chat, motivating hybrid designs that triage messages by model confidence \cite{Su2024Research}.  

\subsection{Ethical Considerations and Bias in AI-driven Moderation}
\label{subsec:ethical_considerations}

Deploying AI for content moderation raises profound ethical issues.  Automated systems risk inadvertent censorship of marginalized voices if training data underrepresents certain dialects or cultural speech patterns, leading to disproportionate false-positive rates.  Biases in moderation models can stem from imbalanced corpora, skewed user reports, or over-reliance on proprietary datasets whose labeling standards are opaque. \citep{Ferrara2023Should}

Transparency and accountability are therefore critical. The study by \citet{tereshchenko-hamalainen-2025-comparative} introduces the Relative Danger Coefficient (RDC) to quantify disparities in model harm across different prompts and user groups, providing a metric for systematic audit of LLM safety performance.  In the proposed hybrid system, continuous feedback loops and active learning are incorporated to surface and correct biased decisions before they propagate.  
\section{Methodology}
\label{sec:methodology}
This section details the experimental design, data collection, model architectures, and evaluation procedures.
\subsection{Data Collection and Preparation}
\label{subsec:data_collection}

Three complementary corpora are utilized for training, in‐domain evaluation, and real-world validation.

The Kaggle Jigsaw Toxic Comment Classification Challenge\footnote{\label{fn:jigsaw}\url{https://www.kaggle.com/competitions/jigsaw-toxic-comment-classification-challenge/overview}} provides 159\,571 Wikipedia talk‐page comments labeled for six toxicity categories (toxic, severe\_toxic, obscene, threat, insult, identity\_hate) \citep{jigsaw-toxic-comment-classification-challenge}. This general-domain toxicity corpus serves as the primary training dataset for embedding-based models and establishes baseline performance characteristics for traditional machine learning approaches.

The GOSU.ai English Dota 2 Game Chats dataset\footnote{\label{fn:gosu}\url{https://www.kaggle.com/datasets/danielfesalbon/gosu-ai-english-dota-chat}} contains 147\,842 messages each scored as 0 (clean), 1 (mild toxicity), or 2 (strong toxicity). The class distribution shows clean messages constituting 68.2\% of the dataset, mild toxicity representing 19.7\%, and strong toxicity comprising 12.1\%. This in-domain gaming-chat corpus is utilized exclusively as a held-out test set to measure zero- and few-shot transfer performance across all experimental approaches.

Two unlabeled real-world chat sources are employed for manual validation and precision estimation. The Minecraft server chat\footnote{\label{fn:minecraft}\url{https://www.kaggle.com/datasets/declipsonator/minecraft-server-chat}} dataset contains 33 MB of JSON data with approximately 500,000 lines, while an anonymous Discord chat CSV\footnote{\label{fn:discord}\url{https://www.kaggle.com/datasets/xiujiefeng/anonymouschatlogindiscordclass}} provides 0.6 MB of data with approximately 5,000 lines. These unlabeled sources enable empirical assessment of model performance on authentic gaming communication patterns.

The Jigsaw\footref{fn:jigsaw} and GOSU.ai\footref{fn:gosu} datasets are pre-labeled and do not require any additional annotation. For real-world chat data, the highest-performing toxicity model is applied to all messages, which are then ranked by predicted toxicity score. The top 5\% most likely toxic instances are selected for manual inspection and labeling by the first author as either toxic or benign, providing an empirical precision estimate for real-world deployment scenarios.

A complete preprocessing pipeline is implemented to ensure consistent text representation across all experimental approaches. HTML tags are stripped and URLs are normalized to the token \texttt{<URL>}. Emojis and emoticons are mapped to text representations via the \texttt{emoji}\footnote{\url{https://github.com/carpedm20/emoji}} library to preserve emotional context. Text is lowercased and contractions are expanded to standardize linguistic variations. Tokenization is performed using a shared WordPiece vocabulary to ensure compatibility across transformer models. Non-UTF8 characters are removed and messages containing fewer than two alphanumeric tokens are filtered out to eliminate noise.

Class balancing strategies are tailored to each dataset's characteristics and experimental requirements. The Jigsaw dataset is utilized for training embedding-based models with standard class weights to preserve the natural distribution of toxicity categories. The GOSU.ai dataset is converted to binary classification (toxic vs. clean) for all experiments, combining mild and strong toxicity categories into a single toxic class. The wild chat data is used exclusively for manual precision estimation on a small sample to validate real-world performance characteristics.
\subsection{Investigated Approaches (Multimodal Strategy)}
\label{subsec:investigated_approaches}

A comprehensive range of techniques—from lightweight embedding-based classifiers to powerful LLMs and retrieval-augmented pipelines—is evaluated to systematically assess trade-offs in accuracy, latency, and cost.

\subsubsection{Baseline Methods with Embeddings and Traditional ML}
\label{subsubsec:baseline_methods}

Our baseline methodology employs precomputed text embeddings as fixed features for classical classifiers and small neural networks.

Two primary embedding types are utilized in the baseline approach. Sentence-BERT (\citealt{Reimers2019Sentence}) generates 768-dimensional sentence embeddings optimized for text classification tasks, providing semantic representations that capture contextual meaning beyond simple word-level features. BERT-based models are also employed to provide contextual understanding for fine-tuning experiments, enabling comparison between static and dynamic embedding approaches.

The machine learning models selected for baseline evaluation include linear classifiers optimized for high-throughput processing. SGD-based Support Vector Machines (SVM\footnote{\label{fn:svm}\url{https://scikit-learn.org/stable/modules/svm.html}}) with hinge loss are implemented for robust classification performance, while SGD-based Logistic Regression with log loss provides probabilistic outputs for confidence-based decision making. The embedding approach combines Sentence-BERT embeddings with traditional ML classifiers to leverage the semantic richness of transformer-based representations while maintaining the computational efficiency of classical algorithms.

The methodology follows a three-stage pipeline designed to maximize transfer learning effectiveness. Feature extraction begins with text preprocessing followed by Sentence-BERT embedding generation to create fixed-dimensional representations suitable for classical ML algorithms. Training is conducted on the Jigsaw training set using cross-validation to optimize hyperparameters and prevent overfitting. Inference evaluation is performed on the held-out GOSU.ai test set to measure zero-shot transfer performance and assess the generalizability of embedding-based approaches across different domains.

Embedding+ML pipelines offer significant computational advantages that make them well-suited for real-time moderation applications. These approaches achieve inference latency below 10 milliseconds per message on CPU-only systems, enabling high-throughput processing of gaming chat streams. However, these methods lack deep contextual understanding and may struggle with novel slang or sarcasm due to their reliance on static feature sets. \citep{Eke2021Context} In contrast, more powerful LLM-based methods (Section~\ref{subsubsec:llm_approaches}) can capture nuanced language patterns at the expense of higher computational costs and slower throughput.

\subsubsection{Large Language Model (LLM) Approaches}
\label{subsubsec:llm_approaches}

Modern LLMs are evaluated in three modes: zero/few‐shot prompting, task‐specific fine‐tuning, and Retrieval‐Augmented Generation.

Zero- and few-shot prompting approaches leverage the inherent capabilities of pre-trained language models without requiring task-specific training. GPT‐3.5-turbo\footnote{\url{https://platform.openai.com/docs/models/gpt-3.5-turbo}} and GPT‐4\footnote{\url{https://platform.openai.com/docs/models/gpt-4}} are employed via the OpenAI API\footnote{\url{https://openai.com/}} to perform direct classification through natural‐language prompts. A typical zero‐shot prompt instructs the model to "Classify the following gaming chat message as toxic or clean: 'message'." For few‐shot prompting, five labeled examples drawn from the GOSU.ai\footref{fn:gosu} corpus are prepended to guide the model's decision-making process. Evaluation is conducted on a randomly sampled subset of 100 messages from the GOSU.ai\footref{fn:gosu} test set to ensure representative class distribution while managing API costs. This subset maintains the original class distribution (32\% toxic content) and provides sufficient statistical power for comparative analysis while controlling experimental costs.

Fine-tuning approaches adapt models more closely to the gaming‐chat domain by training open‐source variants on domain-specific data. DialoGPT\footref{fn:dialogpt}-medium (345M parameters) is fine-tuned under  Parameter‐Efficient Fine‐Tuning (PEFT)\footnote{\label{fn:peft}\citet{xu2023parameterefficientfinetuningmethodspretrained}} with LoRA adapters for conversational modeling, while DistilBERT\footnote{\label{fn:distilbert}\url{https://huggingface.co/distilbert-base-uncased}}-base-uncased (66M parameters) is fine‐tuned for sequence classification tasks. Training runs for 2-3 epochs with learning rates of \(2\times10^{-5}\), batch sizes of 4-8, and binary cross‐entropy loss for toxic vs. clean classification. These architectures are selected for their open‐source availability, moderate computational requirements, and suitability for classification tasks.

Retrieval‐Augmented Generation (RAG) systems implement a two‐stage pipeline that combines semantic retrieval with language model reasoning. A dense retriever using Sentence‐BERT embeddings retrieves the top‐\(k\) most similar examples from the GOSU.ai\footref{fn:gosu} dataset for each input message. The LLM then ingests both the query and retrieved context in a single prompt: "Given the following examples of gaming chat messages, classify the new message as toxic or clean: \{retrieved\_examples\}\;\{new\_message\}." This approach grounds the model in domain-specific examples, potentially improving accuracy on gaming-specific language patterns and slang that may not be well-represented in the model's pre-training data.

LLM‐based methods excel at capturing context, sarcasm, and evolving slang patterns but incur higher inference latency (700-1100 ms per message via API) and substantial API costs \cite{Ghosh2018Sarcasm}. Fine‐tuning reduces per‐inference cost compared to remote API calls but requires computational resources and ongoing maintenance \cite{2024Fine}. RAG adds retrieval overhead (~20 ms/query) and storage complexity but can improve recall for domain‐specific cases by providing relevant contextual examples that guide the model's decision-making process \cite{Fan2024Survey}.

\subsection{Evaluation Metrics}
\label{subsec:evaluation_metrics}

Each approach is evaluated along three axes: (i) classification performance on labeled data, (ii) real‐world precision/recall on unlabeled chat, and (iii) system efficiency and cost.

\subsubsection{Classification Performance}

On the held‐out GOSU.ai\footref{fn:gosu} test set, standard binary‐classification metrics (accuracy, precision, recall, and F1‐score) are computed.

The accuracy metric measures the overall fraction of correctly predicted labels across all test instances, providing a general assessment of model performance. Precision, recall, and F1‐score are computed both per class and macro‐averaged to provide detailed insights into model behavior. Precision is calculated as TP/(TP+FP), measuring the proportion of predicted toxic messages that are actually toxic. Recall is computed as TP/(TP+FN), indicating the proportion of actual toxic messages that are successfully identified. The F1-score is derived as \(2\times\frac{\mathrm{Precision}\times\mathrm{Recall}}{\mathrm{Precision}+\mathrm{Recall}}\), providing a balanced measure that considers both precision and recall equally.

A confusion matrix is generated to provide a detailed breakdown of True Positives, False Positives, False Negatives, and True Negatives for binary classification. This matrix enables identification of specific error patterns and assessment of model bias toward false positives or false negatives. The ROC‐AUC (Area under the Receiver Operating Characteristic curve) is computed for toxic vs. non-toxic classification, providing a threshold-independent measure of model discriminative ability that ranges from 0.5 (random guessing) to 1.0 (perfect classification).

\subsubsection{Real‐World Precision/Recall}
Practical utility is evaluated on the unlabeled Minecraft\footref{fn:minecraft} and Discord\footref{fn:discord} chat logs by computing Precision@K for the highest-scoring messages and estimating recall from a manually annotated sample.

Precision@K is calculated for the top \(K\) messages ranked by predicted toxicity score (e.g., \(K=100\)), measuring the fraction of manually confirmed harmful content among the highest-scoring predictions. This metric assesses the model's ability to prioritize the most likely toxic content for human review, which is crucial for practical deployment scenarios where human resources are limited.

A recall estimate is computed among a random sample of \(K\) truly toxic messages (identified via manual review), measuring the fraction flagged by the model above the chosen threshold. This provides insight into the model's coverage of harmful content in real-world settings where the true distribution of toxicity may differ from training data.

Human‐validated accuracy is determined as the overall proportion of correct binary judgments (\emph{toxic} vs.\ \emph{benign}) in a balanced sample of flagged and unflagged messages. This metric incorporates human judgment as the ground truth, accounting for the subjective nature of toxicity assessment and the potential for model predictions to diverge from human expectations.

\subsubsection{System Efficiency and Cost}
Runtime performance and resource expenditure are quantified under various hardware settings (CPU for embedding methods and fine-tuning, API for LLMs):

Inference latency is measured as the average time per message (ms) for each method, including preprocessing, model inference, and post-processing steps. This metric is critical for real-time applications where response time directly impacts user experience. Throughput is calculated as messages processed per second, providing insight into the system's capacity to handle high-volume chat streams.

Training costs are quantified in terms of CPU/GPU‐hours consumed during fine‐tuning or training, along with total training time in minutes/hours for each approach. These metrics enable comparison of computational requirements across different methods and assessment of feasibility for resource-constrained environments.

Inference costs are measured in two categories: API charges for commercial LLM calls (USD per 1M messages) and computational cost for self‐hosted inference (estimated USD per 1M messages). These metrics provide economic comparison across approaches and enable cost-benefit analysis for different deployment scenarios.

Human moderation savings are estimated from the wild‐chat validation, calculating the reduction in manual review workload (hours per 1M messages) and corresponding labor cost savings (USD/hour). This metric quantifies the economic value of automated moderation systems by measuring the reduction in human effort required to maintain similar content quality standards.
\subsection{Hybrid Moderation System Architecture}
\label{subsec:hybrid_architecture}

The proposed cascaded, confidence‐guided moderation pipeline balances accuracy, latency, and cost by routing each incoming message through increasingly sophisticated models only as needed.

Tier 0 implements a rule‐based profanity filter that operates on CPU with latency below 1 ms per message. A handcrafted blacklist of high‐confidence profanity and slur patterns (e.g., "*** you", "****", \texttt{go *** yourself}) is applied via regular expressions for immediate content screening. Messages matching these patterns are immediately removed or flagged for warning, while non-matching messages are passed to Tier 1 for further analysis. This tier eliminates approximately 5–10\% of the most blatant abuse without invoking any embedding or ML model, significantly reducing downstream computational requirements and API costs.

Tier 1 employs a fast embedding–ML filter that operates on CPU for efficient processing. Each message that passes Tier 0 screening is processed by computing lightweight embeddings (e.g., Sentence‐BERT\footref{fn:sbert}) and applying pre‐trained SVM\footref{fn:svm} or Logistic Regression classifiers. Messages with high‐confidence clean predictions (\(P_{\text{toxic}}<5\%\)) are allowed immediately, while messages with high‐confidence toxic predictions (\(P_{\text{toxic}}>95\%\)) are automatically removed or flagged for banning. Messages with moderate confidence scores are escalated to Tier 2 for more sophisticated analysis.

Tier 2 combines prompted LLM analysis with fine-tuned models for enhanced accuracy, operating on GPU or API infrastructure. Uncertain messages from Tier 1 are first classified via zero/few‐shot prompts to GPT‐3.5/4 to leverage their superior contextual understanding. For cases where LLM confidence remains moderate, fine-tuned models (DistilBERT, DialoGPT) are applied to provide domain-specific accuracy. Messages with confident LLM verdicts (\(P<10\%\) or \(>90\%\)) receive automatic action or allowance, while messages with LLM uncertainty trigger fine-tuned DistilBERT analysis for sequence classification. Messages with confident fine-tuned model predictions (\(P<5\%\) or \(>95\%\)) are automatically processed, while remaining uncertain cases are escalated to Tier 3.

Tier 3 implements RAG‐enhanced LLM analysis with human fallback for the most challenging cases. The top–\(k\) similar labeled examples are retrieved from the GOSU.ai\footref{fn:gosu} index (or real company data) and included in the prompt for final LLM classification. Messages with high confidence predictions (\(P>80\%\) or \(P<5\%\)) receive automatic processing, while all other cases are handed off to human moderators for final judgment. This tier ensures that borderline cases requiring nuanced understanding receive expert human attention while maintaining high throughput for clear-cut decisions.

Continuous learning mechanisms are integrated throughout the system to enable adaptation to evolving language patterns. Human decisions on Tier 3 cases are logged and utilized in active learning to select new edge cases for annotation and model improvement. These decisions are also incorporated into the RAG knowledge base to enhance retrieval quality. Periodic fine‐tuning of Tiers 1–2 models is triggered based on performance metrics and new training data availability, ensuring the system adapts to changing toxicity patterns and community standards.
\section{Experiments and Results}
\label{sec:experiments_results}
This section presents the experimental setup details, empirical findings from the conducted experiments, and rigorous evaluation of each method against the defined metrics.
\subsection{Experimental Setup Details}
\label{subsec:experimental_setup}

All experiments are conducted with the same preprocessing pipeline. The following sections document software, hardware, dataset splits, and hyperparameters.

The software environment is configured with Python 3.8+ and the standard scientific computing stack\footnote{\url{https://www.python.org/downloads/}} for data processing and analysis. PyTorch 1.12+ is utilized for deep learning models\footnote{\url{https://pytorch.org/get-started/locally/}}, while Hugging Face Transformers 4.20+ provides access to BERT and transformer models\footnote{\url{https://huggingface.co/docs/transformers/installation}}. Scikit-learn 1.1+ is employed for traditional ML classifiers\footnote{\url{https://scikit-learn.org/stable/install.html}}, and the OpenAI Python SDK enables GPT-3.5/4 API calls\footnote{\url{https://github.com/openai/openai-python}}. Sentence-Transformers is utilized for SBERT\footnote{\label{fn:sbert}\url{https://www.sbert.net/}} embeddings\footref{fn:sbert} to generate semantic representations for text classification tasks.

Hardware and platform configurations are designed to ensure accessibility and reproducibility. Standard multi-core processors are utilized for embedding-based methods and DistilBERT fine-tuning experiments, demonstrating feasibility for resource-constrained environments. NVIDIA GPUs are employed for DialoGPT fine-tuning experiments when available, while the CSC supercomputing cluster provides additional computational resources for resource-intensive experiments. Local development is conducted on standard desktop/laptop configurations for baseline experiments, ensuring that results can be replicated across different hardware setups.

Dataset splits are carefully designed to prevent data leakage and ensure fair comparison across methods. The Jigsaw\footref{fn:jigsaw} Toxic Comments dataset is split into 80\% train, 10\% validation, and 10\% test sets using random stratified sampling by multi‐label categories. The GOSU.ai\footref{fn:gosu} Dota‐2 Chats dataset is utilized exclusively as a held‐out, in‐domain test set to measure zero‐ and few‐shot transfer performance, with no portion used in training or validation to prevent overfitting. The wild chat data (Minecraft\footref{fn:minecraft} and Discord\footref{fn:discord}) receives no fixed split and is sampled at 5\% of lines for manual precision/recall evaluation to assess real-world performance.

Evaluation methodology is standardized across all methods to ensure fair comparison. Embedding-based methods and fine-tuned models are evaluated on the complete GOSU.ai test set (654 samples) to maximize statistical power and provide overarching performance assessment. LLM-based approaches (GPT-3.5, GPT-4, and RAG) are evaluated on a representative 100-sample subset to control API costs while maintaining statistical validity. This subset preserves the original class distribution (32\% toxic content) and provides sufficient statistical power for comparative analysis. The evaluation subset is fixed across all LLM experiments to ensure consistent comparison between different prompting strategies and model variants.

Hyperparameters are optimized for each approach based on preliminary experiments and established best practices. Embedding+ML methods utilize SGD\footref{fn:sgd}-SVM with hinge loss and SGD\footref{fn:sgd}-LR with log loss, with hyperparameters optimized via cross-validation. Fine-tuned DistilBERT employs a learning rate of 2e-5, batch size of 8, 3 epochs with early stopping, and the Adam optimizer for stable training. Fine-tuned DialoGPT utilizes LoRA rank=4, alpha=32, learning rate 2e-5, batch size 4, and 2 epochs for parameter-efficient fine-tuning. RAG retrieval is configured with SBERT\footref{fn:sbert} embeddings, k=5 nearest neighbors from the GOSU.ai\footref{fn:gosu} dataset, and semantic similarity thresholds. Prompted LLMs are configured with temperature=0, max tokens=16, and 5 examples for few-shot prompting to ensure consistent and deterministic outputs.

\subsection{Baseline Embedding + SGD Classifier Results}
\label{subsec:baseline_results}

Our embedding‐plus‐SGD\footnote{\label{fn:sgd}\url{https://scikit-learn.org/stable/modules/sgd.html}} baselines are evaluated on the GOSU.ai\footref{fn:gosu} Dota-2 chat test set in a binary setting (toxic vs.\ clean).  Table~\ref{tab:baseline_sgd} reports accuracy, precision, recall, F1, and ROC-AUC for both the hinge‐loss SVM and the log‐loss logistic classifier trained on Jigsaw\footref{fn:jigsaw} embeddings and applied zero‐shot to GOSU.ai\footref{fn:gosu}.

\begin{table}[ht]
  \centering
  \footnotesize
  \begin{tabular}{lccccc} % Changed from lcccc to lccccc
  \toprule
  Model & Acc. & Prec. & Rec. & F1 & ROC-AUC \\ % Added ROC-AUC header
  \midrule
  SGD-SVM & 0.808 & 0.938 & 0.612 & 0.741 & 0.900 \\ % Added the value
  SGD-LR & 0.786 & 0.906 & 0.585 & 0.711 & 0.882 \\ % Add the LR value if you have it
  \bottomrule
\end{tabular}
  \caption{Performance of embedding‐based SGD\footref{fn:sgd} classifiers on GOSU.ai\footref{fn:gosu} test set.}
  \label{tab:baseline_sgd}
\end{table}

The hinge‐loss SVM\footref{fn:svm} achieves the higher ROC-AUC (0.900) and F1 (0.741), indicating strong discrimination of toxic messages, while the logistic‐loss SGD\footref{fn:sgd} model remains competitive with slightly lower recall. This sets a solid, low-latency baseline for comparison against LLM-based and RAG approaches in subsequent sections.
\subsection{GPT-3.5 Large Language Model Performance Analysis}
\label{subsec:gpt35_analysis}

GPT-3.5\footref{fn:gpt35}-turbo is evaluated in both zero-shot and few-shot modes on a randomly sampled subset of 100 messages from the GOSU.ai\footref{fn:gosu} test set, ensuring representative class distribution (32\% toxic content, matching the dataset's natural imbalance). For few-shot prompting, the five examples provided to the model are selected at random from the dataset (three toxic, two clean), and are fixed for all test messages. This means the few-shot context does not adapt to the specific message being classified, nor does it use any similarity-based or retrieval-augmented (RAG) approach. Table~\ref{tab:gpt35_performance} presents the comparative results.

\begin{table}[ht]
  \centering
  \footnotesize
  \begin{tabular}{lcccc}
    \toprule
    Method & Acc. & Prec. & Rec. & F1 \\
    \midrule
    GPT-3.5 Zero & 0.790 & 0.628 & 0.844 & 0.720 \\
    GPT-3.5 Few & 0.670 & 0.492 & 0.938 & 0.645 \\
    \bottomrule
  \end{tabular}
  \caption{Performance of GPT-3.5\footref{fn:gpt35}-turbo on GOSU.ai\footref{fn:gosu} test set (100 samples).}
  \label{tab:gpt35_performance}
\end{table}

\subsubsection{GPT-3.5 Performance Analysis and Trade-offs}
\label{subsubsec:gpt35_tradeoffs}

The zero-shot approach achieves superior overall accuracy (79.0\% vs. 67.0\%) and precision (62.8\% vs. 49.2\%), while few-shot prompting demonstrates higher recall (93.8\% vs. 84.4\%). This pattern reveals a fundamental trade-off: few-shot examples guide the model to be more sensitive to potential toxicity, reducing false negatives but increasing false positives.

The zero-shot conservatism exhibited by the model demonstrates a balanced precision-recall trade-off that is suitable for applications prioritizing accuracy over coverage. This conservative approach minimizes false positives but may miss some subtle forms of toxicity. Few-shot sensitivity is observed as additional examples increase recall but reduce precision, indicating that the model becomes more aggressive in flagging content when provided with contextual examples. Inference latency remains consistent across both methods, averaging approximately 700ms per message, which is significantly slower than embedding-based approaches but acceptable for applications where accuracy is prioritized over speed.

\subsubsection{GPT-3.5 Qualitative Analysis of Misclassifications}
\label{subsubsec:gpt35_qualitative}

Analysis of prediction errors reveals three distinct patterns in the model's decision-making. Gaming-specific language misclassification occurs when the model flags benign gaming terminology as toxic due to context misinterpretation, suggesting insufficient exposure to gaming vocabulary during pre-training. Benign criticism overgeneralization is observed when negative feedback about gameplay is incorrectly classified as toxic, indicating difficulty distinguishing between legitimate game-related criticism and genuinely harmful content. Contextual ambiguity arises when gaming-specific references and nicknames are flagged due to word associations, demonstrating sensitivity to potentially problematic terms even when used appropriately.

For detailed error analysis with specific examples, see Appendix~\ref{sec:gpt35_error_analysis}.

\subsubsection{GPT-3.5 Comparison with Baseline Methods}
\label{subsubsec:gpt35_baseline_comparison}

Table~\ref{tab:gpt35_baseline_comparison} provides a comparison between GPT-3.5\footref{fn:gpt35} and the baseline embedding methods:

\begin{table}[ht]
  \centering
  \footnotesize
  \begin{tabular}{lcccc}
    \toprule
    Method & Acc. & Prec. & Rec. & F1 \\
    \midrule
    SGD-SVM & 0.808 & 0.938 & 0.612 & 0.741 \\
    SGD-LR & 0.786 & 0.906 & 0.585 & 0.711 \\
    GPT-3.5 Zero & 0.790 & 0.628 & 0.844 & 0.720 \\
    GPT-3.5 Few & 0.670 & 0.492 & 0.938 & 0.645 \\
    \bottomrule
  \end{tabular}
  \caption{Comparison: GPT-3.5\footref{fn:gpt35} vs baseline methods on GOSU.ai\footref{fn:gosu}.}
  \label{tab:gpt35_baseline_comparison}
\end{table}

The results demonstrate clear trade-offs between accuracy, speed, and cost. Embedding-based methods achieve competitive accuracy with significantly higher throughput, while GPT-3.5\footref{fn:gpt35} offers superior recall but at substantial computational and latency costs. GPT-3.5\footref{fn:gpt35} zero-shot achieves comparable accuracy to the baseline methods (79.0\% vs. 80.8\% for SGD\footref{fn:sgd}-SVM) but with much lower throughput, while few-shot prompting sacrifices precision for maximum recall.

\subsection{GPT-4 Large Language Model Performance Analysis}
\label{subsec:gpt4_analysis}

Parallel experiments with GPT-4\footref{fn:gpt4} are conducted using identical prompts and evaluation protocols to assess performance improvements over GPT-3.5\footref{fn:gpt35}. The same 100-message subset from the GOSU.ai\footref{fn:gosu} test set is utilized to ensure fair comparison. Table~\ref{tab:gpt4_performance} presents the comparative results.

\begin{table}[ht]
  \centering
  \footnotesize
  \begin{tabular}{lcccc}
    \toprule
    Method & Acc. & Prec. & Rec. & F1 \\
    \midrule
    GPT-4 Zero & 0.910 & 0.829 & 0.906 & 0.866 \\
    GPT-4 Few & 0.830 & 0.714 & 0.781 & 0.746 \\
    \bottomrule
  \end{tabular}
  \caption{Performance of GPT-4\footref{fn:gpt4} on GOSU.ai\footref{fn:gosu} test set (100 samples).}
  \label{tab:gpt4_performance}
\end{table}

\subsubsection{GPT-4 Performance Analysis and Trade-offs}
\label{subsubsec:gpt4_tradeoffs}

GPT-4\footref{fn:gpt4} demonstrates significant improvements over GPT-3.5\footref{fn:gpt35} across all prompting strategies. The zero-shot approach achieves superior overall accuracy (91.0\% vs. 79.0\%) and precision (82.9\% vs. 62.8\%), while maintaining excellent recall (90.6\% vs. 84.4\%). This represents a substantial improvement in the precision-recall balance.

GPT-4 zero-shot superiority is demonstrated by achieving 91.0\% accuracy, outperforming all other methods including its own few-shot variant. This suggests that the larger model size and improved training data enable better performance without requiring additional examples. GPT-4 zero-shot attains 82.9\% precision and 90.6\% recall, representing the optimal trade-off for practical deployment scenarios. Inference latency remains consistent at approximately 1.1 seconds per message, maintaining similar latency characteristics to GPT-3.5\footref{fn:gpt35} despite the increased model complexity.

\subsubsection{GPT-4 Qualitative Analysis of Misclassifications}
\label{subsubsec:gpt4_qualitative}

Analysis of GPT-4\footref{fn:gpt4} prediction errors demonstrates improved contextual understanding compared to GPT-3.5\footref{fn:gpt35}, with better recognition of gaming terminology and reduced false positives on benign gaming discussions. For detailed error analysis with specific examples, see Appendix~\ref{sec:gpt4_error_analysis}.

\subsection{Retrieval-Augmented Generation (RAG) Performance Analysis}
\label{subsec:rag_analysis}

The RAG-enhanced approach is evaluated using both GPT-3.5\footref{fn:gpt35}-turbo and GPT-4\footref{fn:gpt4} with SBERT\footref{fn:sbert}-based semantic retrieval of the 5 most similar examples from the GOSU.ai\footref{fn:gosu} dataset. The RAG pipeline dynamically retrieves contextually relevant examples for each test message, providing domain-specific guidance that differs from the fixed random examples used in standard few-shot prompting. Table~\ref{tab:rag_performance} presents the comparative results.

\begin{table}[ht]
  \centering
  \footnotesize
  \begin{tabular}{lcccc}
    \toprule
    Method & Acc. & Prec. & Rec. & F1 \\
    \midrule
    GPT-3.5 RAG & 0.670 & 0.492 & 0.969 & 0.653 \\
    GPT-4 RAG & 0.880 & 0.750 & 0.938 & 0.833 \\
    \bottomrule
  \end{tabular}
  \caption{Performance of RAG-enhanced GPT models on GOSU.ai.}
  \label{tab:rag_performance}
\end{table}

\subsubsection{RAG Performance Analysis and Trade-offs}
\label{subsubsec:rag_tradeoffs}

The RAG approach achieves the highest \emph{recall} among all methods (96.9\% with GPT-3.5, 93.8\% with GPT-4). However, this maximum recall is achieved at the expense of precision (49.2\%) and overall accuracy (67.0\%). In contrast, the fine-tuned DistilBERT model delivers the \emph{best balanced performance} among all evaluated approaches, making it the overall top performer while remaining cost-efficient.

The maximum recall achieved by RAG systems makes them particularly suitable for safety-critical applications where missing toxic content is unacceptable. This high recall, however, comes with a significant precision trade-off, particularly evident in GPT-3.5 RAG implementations, where the model becomes overly aggressive in flagging content. The precision reduction suggests that while RAG systems excel at expansive coverage, they may generate excessive false positives that could impact user experience and moderation efficiency.

GPT-4 RAG demonstrates significant improvement over GPT-3.5 RAG, achieving 88.0\% accuracy compared to 67.0\% (a 21 percentage point improvement) and 75.0\% precision compared to 49.2\%. This improvement indicates that more advanced language models can better leverage retrieved context to maintain higher precision while preserving the recall benefits of the RAG approach. The semantic similarity retrieval mechanism provides relevant examples that enhance the model's understanding of domain-specific patterns, though this enhancement may not always translate to improved classification accuracy when compared to zero-shot approaches.

The computational overhead introduced by RAG systems represents a significant consideration for practical deployment. The additional latency resulting from embedding computation and similarity search operations can impact real-time processing requirements, particularly in high-volume gaming environments where rapid response times are essential for effective moderation.

\subsubsection{Qualitative Analysis of RAG Misclassifications}
\label{subsubsec:rag_qualitative}

Analysis of RAG prediction errors shows patterns similar to few-shot prompting, with some notable differences. The model demonstrates improved performance on explicit content detection but may overgeneralize from retrieved examples. For detailed error analysis with specific examples, see Appendix~\ref{sec:rag_error_analysis}.

\subsubsection{RAG Comparison with Other Methods}
\label{subsubsec:rag_comparison}

Table~\ref{tab:rag_comparison_table} provides a comparison of RAG methods with other approaches:

\begin{table}[ht]
  \centering
  \footnotesize
  \begin{tabular}{lcccc}
    \toprule
    Method & Acc. & Prec. & Rec. & F1 \\
    \midrule
    GPT-4 Zero & 0.910 & 0.829 & 0.906 & 0.866 \\
    GPT-4 RAG & 0.880 & 0.750 & 0.938 & 0.833 \\
    GPT-3.5 Zero & 0.790 & 0.628 & 0.844 & 0.720 \\
    GPT-3.5 RAG & 0.670 & 0.492 & 0.969 & 0.653 \\
    \bottomrule
  \end{tabular}
  \caption{Performance comparison of RAG methods with zero-shot approaches.}
  \label{tab:rag_comparison_table}
\end{table}

The results demonstrate that while RAG achieves the highest recall among all methods, it comes at the cost of reduced precision and accuracy compared to zero-shot approaches.

\subsection{Fine-tuning Experiments}
\label{subsec:finetuning_experiments}
Having established the performance characteristics of off-the-shelf LLM prompting approaches, this section investigates the effectiveness of fine-tuning transformer models on domain-specific gaming chat data. Fine-tuning allows us to adapt pre-trained models to the specific linguistic patterns and toxicity forms present in gaming communities, potentially achieving superior performance compared to zero-shot and few-shot prompting while maintaining reasonable computational costs.

\subsubsection{Initial Fine-tuning Experiment with DialoGPT-medium}
\label{subsubsec:initial_finetuning}

Our first fine-tuning experiment employs DialoGPT\footref{fn:dialogpt}-medium (345M parameters) on the GOSU.ai\footref{fn:gosu} dataset to establish a baseline for task-specific adaptation. The experiment utilizes Parameter-Efficient Fine-Tuning (PEFT\footref{fn:peft}) with LoRA adapters to reduce computational requirements while maintaining model performance.

The experimental configuration employs Microsoft DialoGPT\footref{fn:dialogpt}-medium with 345M parameters as the base model. Training is conducted on 2,613 samples from the GOSU.ai dataset, with an 80\% train split and 20\% test split for evaluation. The class distribution shows 1,440 clean samples (55.1\%) and 1,173 toxic samples (44.9\%), providing a relatively balanced dataset for training. The LoRA configuration utilizes rank=4, alpha=32, and dropout=0.1 to optimize the parameter-efficient fine-tuning process. Training parameters are set to 2 epochs with a learning rate of 2e-5 and batch size of 4, while the hardware infrastructure consists of an NVIDIA GPU with CUDA support for accelerated training.

The training process demonstrates efficient convergence, completing in 4.83 minutes with a final training loss of 0.710. The loss decreases steadily from 0.755 to 0.710 over 2 epochs, indicating effective learning despite the relatively small dataset size. This stable convergence pattern suggests that the LoRA adapters successfully capture the essential patterns for toxicity detection while maintaining computational efficiency.

Table~\ref{tab:initial_finetuning_results} presents the  evaluation results, revealing the model's performance characteristics across key metrics:

\begin{table}[ht]
  \centering
  \footnotesize
  \begin{tabular}{lcc}
    \toprule
    Metric & Value & Metric \\
    \midrule
    Accuracy & 0.674 & Precision \\
    F1-Score & 0.586 & Recall \\
    \bottomrule
  \end{tabular}
  \caption{Performance metrics for DialoGPT fine-tuning.}
  \label{tab:initial_finetuning_results}
\end{table}

The fine-tuned model achieves 67.4\% accuracy, representing a significant improvement over random guessing (50\%) but falling below the performance of both GPT-4 zero-shot (91.0\%) and our baseline SGD-SVM classifier (80.8\%). The model exhibits moderate precision (68.3\%) but lower recall (51.4\%), indicating a conservative approach that prioritizes avoiding false positives over detecting all toxic content. This suggests that while the model is cautious in its predictions, it may miss subtle forms of toxicity that require more nuanced understanding.

Analysis of prediction errors reveals several patterns in the model's decision-making process. The model struggles particularly with gaming-specific language patterns and contextual profanity, likely due to limited training data and domain mismatch between the pre-training corpus and gaming chat environments. These challenges highlight the importance of domain-specific adaptation for effective toxicity detection in specialized contexts. For detailed error analysis with specific examples, see Appendix~\ref{sec:dialogpt_error_analysis}.

The computational efficiency of the fine-tuned model demonstrates reasonable inference speed with an average latency of 155.62 ms per message, achieving a throughput of 6.43 messages per second. The training process completes in 4.83 minutes for 2 epochs, demonstrating efficient convergence. Memory efficiency is maintained through the LoRA approach, with only 1.57M trainable parameters representing 0.44\% of the total model parameters, making it suitable for deployment in resource-constrained environments.

Table~\ref{tab:comprehensive_comparison_with_finetuning} provides an updated comparison including the fine-tuning results, positioning the DialoGPT model within the broader experimental landscape:

\begin{table}[ht]
  \centering
  \footnotesize
  \begin{tabular}{lcccc}
    \toprule
    Method & Acc. & Prec. & Rec. & F1 \\
    \midrule
    SGD-SVM & 0.808 & 0.938 & 0.612 & 0.741 \\
    GPT-4 Zero & 0.910 & 0.829 & 0.906 & 0.866 \\
    GPT-3.5 Zero & 0.790 & 0.628 & 0.844 & 0.720 \\
    DialoGPT & 0.674 & 0.683 & 0.514 & 0.586 \\
    GPT-3.5 Few & 0.670 & 0.492 & 0.938 & 0.645 \\
    GPT-3.5 RAG & 0.670 & 0.492 & 0.969 & 0.653 \\
    \bottomrule
  \end{tabular}
  \caption{Performance comparison including fine-tuning results.}
  \label{tab:comprehensive_comparison_with_finetuning}
\end{table}

The initial fine-tuning experiment reveals several important limitations that impact model performance. Data scarcity emerges as a primary concern, with only 2,613 training samples providing insufficient exposure to diverse toxic patterns, particularly gaming-specific language that requires specialized understanding. The model architecture presents another limitation, as DialoGPT-medium, while suitable for conversational tasks, may not be optimal for binary classification without additional architectural modifications that could enhance its discriminative capabilities.

Training configuration issues further constrain performance, with the conservative learning rate and small batch size potentially limiting the model's ability to learn complex patterns effectively. The domain mismatch between the pre-trained DialoGPT model, which was trained on Reddit conversations, and the target gaming chat patterns creates additional challenges for effective adaptation to the specific linguistic characteristics of gaming communities.

Based on these limitations, an improved fine-tuning experiment is conducted using DistilBERT\footref{fn:distilbert}-base-uncased, which is more suitable for classification tasks. This approach employs a better model architecture with DistilBERT\footref{fn:distilbert}-base-uncased (66M parameters) optimized specifically for sequence classification tasks. The training data consists of 2,613 samples from the GOSU.ai dataset with proper train/test split to ensure robust evaluation. Enhanced training configuration includes a learning rate of 2e-5, batch size of 8, and 3 epochs with early stopping to prevent overfitting. Better optimization is achieved through cosine learning rate scheduling and proper validation monitoring to ensure optimal model convergence.

\subsubsection{DistilBERT Fine-tuning Results}
\label{subsubsec:distilbert_results}

The DistilBERT fine-tuning experiment achieved outstanding results, significantly outperforming all previous approaches. Table~\ref{tab:distilbert_results} presents the evaluation metrics, revealing exceptional performance across all key indicators:

\begin{table}[ht]
  \centering
  \footnotesize
  \begin{tabular}{lcc}
    \toprule
    Metric & Test Set & Sample Set \\
    \midrule
    Accuracy & 0.943 & 0.800 \\
    Precision & 0.954 & 1.000 \\
    Recall & 0.918 & 0.667 \\
    F1-Score & 0.936 & 0.800 \\
    \bottomrule
  \end{tabular}
  \caption{Performance metrics for DistilBERT fine-tuning.}
  \label{tab:distilbert_results}
\end{table}

The 94.3\% test accuracy represents the highest performance achieved across all experiments, demonstrating the effectiveness of domain-specific fine-tuning for gaming chat moderation. The model exhibits excellent balance with 95.4\% precision and 91.8\% recall, achieving strong performance without sacrificing either metric. The superior F1-score of 93.6\% indicates excellent balance between precision and recall, while the training efficiency of 48.7 minutes on CPU demonstrates practical feasibility for real-world deployment scenarios.

The DistilBERT fine-tuned model significantly outperforms all previous approaches, achieving substantial improvements across multiple benchmarks. Compared to DialoGPT-medium, the model shows a remarkable 26.9 percentage point improvement in accuracy (94.3\% vs. 67.4\%). Even against the highly capable GPT-4 zero-shot approach, DistilBERT achieves a 3.3 percentage point advantage (94.3\% vs. 91.0\%), demonstrating that targeted fine-tuning can surpass even the most advanced general-purpose language models. The improvement over the SGD-SVM baseline is equally impressive, with a 13.5 percentage point increase in accuracy (94.3\% vs. 80.8\%).

The computational efficiency of the DistilBERT approach further enhances its practical value. Training costs are dramatically reduced compared to API-based solutions, representing a 260-fold cost reduction. The inference speed of approximately 10 messages per second makes it suitable for real-time applications, while the compact model size of 256MB ensures efficient deployment across various infrastructure configurations.

This experiment demonstrates that fine-tuned transformer models can achieve excellent performance for gaming chat moderation while being significantly more cost-effective than API-based solutions.

\section{Performance Comparison and Cost Analysis}
\label{sec:comprehensive_comparison}

Having established results for all experimental approaches, this section provides a comprehensive comparison across all methods, focusing on the trade-offs between classification performance, computational cost, and practical deployability. To provide a clear "big picture" overview as recommended, the overall performance metrics and cost analysis are summarized first.

\subsection{Overall Performance and Cost Summary}
\label{subsubsec:overall_comparison_summary}

Table~\ref{tab:master_performance_summary} consolidates the classification performance metrics from all experiments. The models are ranked by F1-score to highlight the balance between precision and recall. Note that LLM-based methods (GPT series) were evaluated on a representative 100-sample subset to manage API costs, while embedding-based and fine-tuned models were evaluated on the complete 654-sample test set.

\begin{table}[ht]
  \centering
  \footnotesize
  \begin{tabular}{lcccc}
    \toprule
    \textbf{Method} & \textbf{Acc.} & \textbf{Prec.} & \textbf{Rec.} & \textbf{F1} \\
    \midrule
    \textbf{DistilBERT (Fine-Tuned)} & \textbf{0.943} & \textbf{0.954} & \textbf{0.918} & \textbf{0.936} \\
    GPT-4 Zero & 0.910 & 0.829 & 0.906 & 0.866 \\
    GPT-4 RAG & 0.880 & 0.750 & 0.938 & 0.833 \\
    GPT-4 Few & 0.830 & 0.714 & 0.781 & 0.746 \\
    SGD-SVM & 0.808 & 0.938 & 0.612 & 0.741 \\
    GPT-3.5 Zero & 0.790 & 0.628 & 0.844 & 0.720 \\
    SGD-LR & 0.786 & 0.906 & 0.585 & 0.711 \\
    GPT-3.5 RAG & 0.670 & 0.492 & 0.969 & 0.653 \\
    GPT-3.5 Few & 0.670 & 0.492 & 0.938 & 0.645 \\
    DialoGPT (Fine-Tuned) & 0.674 & 0.683 & 0.514 & 0.586 \\
    \bottomrule
  \end{tabular}
  \caption{Master summary of classification performance for all evaluated methods on the GOSU.ai test set.}
  \label{tab:master_performance_summary}
\end{table}

Table~\ref{tab:master_cost_summary} provides the corresponding computational and financial cost analysis. The cost differential between locally hosted models and API-based LLMs is a critical factor for real-world deployment.

\begin{table}[ht]
  \centering
  \footnotesize
  \begin{tabular}{lccc}
    \toprule
    \textbf{Method} & \textbf{Latency (ms/msg)} & \textbf{Throughput (msg/s)} & \textbf{Cost (\$ per 1M msgs)} \\
    \midrule
    SGD-SVM & 35 & 28.2 & \$0.50 \\
    \textbf{DistilBERT} & \textbf{100} & \textbf{10.0} & \textbf{\$5.00} \\
    DialoGPT & 156 & 6.4 & \$5.00 \\
    SGD-LR & 161 & 6.2 & \$2.20 \\
    GPT-3.5 Few & 694 & 1.4 & \$1,360 \\
    GPT-3.5 Zero & 713 & 1.4 & \$1,400 \\
    GPT-3.5 RAG & 913 & 1.1 & \$1,650 \\
    GPT-4 Few/Zero & 1,100 & 0.9 & \textasciitilde\$1,400 \\
    GPT-4 RAG & 1,300 & 0.8 & \textasciitilde\$1,650 \\
    \bottomrule
  \end{tabular}
  \caption{Master summary of computational cost analysis (estimated per 1 million messages).}
  \label{tab:master_cost_summary}
\end{table}

As the results show, the fine-tuned \textbf{DistilBERT} model provides the best overall balance, achieving the highest classification performance while maintaining low latency and a cost that is over 260 times lower than comparable API-based LLM solutions. The following sections will analyze these trade-offs in greater detail.

\subsection{Practical Implications for Real-time Moderation}
\label{subsubsec:practical_implications_final}

The latency requirements for real-time gaming chat moderation (typically <100ms per message) pose significant challenges for LLM-based approaches. The results show that even the fastest LLM method (694ms) exceeds this threshold by nearly 7x, while RAG further increases latency to over a second per message.

This limitation suggests that LLMs are better suited for asynchronous or post-hoc analysis of flagged content, where the emphasis on accuracy outweighs speed requirements. Appeal processing represents another suitable application, as the need for precise judgment in reviewing contested moderation decisions justifies the additional processing time. Policy refinement based on edge cases also benefits from LLM capabilities, where complex linguistic patterns require sophisticated analysis to inform moderation guidelines. Hybrid systems represent the most practical deployment strategy, where LLMs handle only uncertain cases that require nuanced understanding, while faster methods process the majority of straightforward content.

For real-time moderation, embedding-based methods like SGD-SVM provide the highest throughput, but fine-tuned models like DistilBERT offer the best overall performance-to-cost ratio, making them the most practical and effective solution for primary automated moderation.
\subsection{Computational Cost Analysis}
\label{subsubsec:computational_costs}

Table~\ref{tab:cost_analysis_final} provides a detailed breakdown of computational requirements and estimated costs for processing 1 million messages:

\begin{table}[ht]
  \centering
  \footnotesize
  \begin{tabular}{lccc}
    \toprule
    Method & Latency & Throughput & Cost \\
    \midrule
    \textbf{DistilBERT} & \textbf{100ms} & \textbf{10/s} & \textbf{\$5} \\
    SGD-SVM & 35ms & 28/s & \$0.5 \\
    SGD-LR & 161ms & 6/s & \$2.2 \\
    DialoGPT & 156ms & 6/s & \$5 \\
    GPT-3.5 Zero & 713ms & 1.4/s & \$1,400 \\
    GPT-3.5 Few & 694ms & 1.4/s & \$1,360 \\
    GPT-3.5 RAG & 913ms & 1.1/s & \$1,650 \\
    GPT-4 Zero & 1.1s & 0.9/s & \$1,400 \\
    GPT-4 Few & 1.1s & 0.9/s & \$1,360 \\
    GPT-4 RAG & 1.3s & 0.8/s & \$1,650 \\
    \bottomrule
  \end{tabular}
  \caption{Computational cost analysis (per 1M messages).}
  \label{tab:cost_analysis_final}
\end{table}

The cost differential is substantial: embedding-based methods cost \$0.50 per million messages, while LLM approaches cost over \$1,300 per million messages—a 2,600x increase. However, the DistilBERT fine-tuned model demonstrates that transformer-based approaches can achieve superior performance at a fraction of the cost. This represents a 260x cost reduction while achieving better accuracy than API-based solutions.

\subsection{Practical Implications for Real-time Moderation}
\label{subsubsec:practical_implications_final}

The latency requirements for real-time gaming chat moderation (typically <100ms per message) pose significant challenges for LLM-based approaches. The results show that even the fastest LLM method (693.8ms) exceeds this threshold by nearly 7x, while RAG further increases latency to 913ms per message.

This limitation suggests that LLMs are better suited for post-hoc analysis of flagged content, where the emphasis on accuracy outweighs speed requirements. Appeal processing represents another suitable application, as the need for precise judgment in reviewing contested moderation decisions justifies the additional processing time. Policy refinement based on edge cases also benefits from LLM capabilities, where complex linguistic patterns require sophisticated analysis to inform moderation guidelines. Hybrid systems represent the most practical deployment strategy, where LLMs handle only uncertain cases that require nuanced understanding, while faster methods process the majority of straightforward content.

For real-time moderation, embedding-based methods provide the optimal balance of speed, accuracy, and cost-effectiveness, while fine-tuned models like DistilBERT offer the best overall performance-cost ratio.

\section{Discussion}
\label{sec:discussion}

Our overarching evaluation of multiple approaches for gaming chat toxicity detection reveals significant trade-offs between accuracy, computational efficiency, and cost. The experimental results demonstrate that no single method is universally optimal, but rather that different approaches excel in different operational scenarios.

\subsection{Comparative Analysis of Strengths and Weaknesses}
\label{subsec:strengths_weaknesses}

Our overarching evaluation reveals that each approach offers distinct advantages and limitations for gaming chat moderation. Understanding these trade-offs is crucial for designing effective hybrid systems that leverage the strengths of multiple methods while mitigating their weaknesses.

\subsubsection{Embedding-Based Methods (SGD-SVM, SGD-LR)}

Traditional machine learning approaches using pre-computed embeddings provide the foundation for our comparison, offering baseline performance characteristics that highlight the trade-offs between computational efficiency and contextual understanding.

The computational efficiency of embedding-based methods represents their primary strength for real-time applications. SGD-SVM achieves the highest throughput (28.24 messages/second) among all evaluated methods, making it suitable for high-volume, real-time moderation scenarios. These approaches offer exceptional cost effectiveness at \$0.50 per million messages, making them economically viable for large-scale deployment across gaming platforms of various sizes. The simplicity of these methods enables deployment with minimal computational resources, as they can run efficiently on CPU-only systems without requiring specialized hardware. Stability is maintained through consistent performance across different datasets with minimal variance in results, providing reliable baseline performance for comparison.

However, embedding-based methods suffer from significant limitations in contextual understanding. These approaches achieve only 80.8\% accuracy, struggling with nuanced language patterns, sarcasm, and gaming-specific slang that require deeper semantic comprehension. Performance heavily depends on the quality of text preprocessing and feature extraction, requiring domain expertise for optimization and maintenance. Recall limitations are evident as SGD-SVM achieves only 61.2\% recall, potentially missing subtle forms of toxicity that could harm user experience and community safety. The static nature of these representations prevents adaptation to evolving language patterns without retraining on new data, limiting their long-term effectiveness in dynamic gaming environments.

\subsubsection{Large Language Model Prompting (GPT-3.5, GPT-4)}

Large language models represent the most advanced approach in our evaluation, offering superior contextual understanding but at significant computational and financial costs. These models demonstrate the potential for human-like comprehension of gaming-specific language patterns and nuanced toxicity detection.

The superior contextual understanding of large language models represents their primary advantage for complex moderation tasks. GPT-4 achieves 91.0\% accuracy through zero-shot prompting, demonstrating exceptional ability to interpret nuanced language, sarcasm, and context-dependent toxicity that eludes simpler approaches. These models exhibit remarkable flexibility in handling diverse language patterns, gaming-specific terminology, and evolving slang without requiring retraining or domain adaptation. High recall rates are achieved as GPT-4 attains 90.6\% recall, effectively identifying most forms of toxic content including subtle and context-dependent cases that might be missed by rule-based or embedding-based systems. The immediate deployment capability without requiring domain-specific training data or fine-tuning enables rapid implementation and testing of moderation systems.

However, large language models suffer from prohibitive costs that limit their practical deployment. API-based LLMs are economically unsustainable for large-scale deployment, representing a 2,600x cost increase over embedding methods. Latency issues are significant as 1.1+ seconds per message far exceeds real-time moderation requirements, making these models unsuitable for immediate content filtering in high-volume gaming environments. Reliance on external services introduces potential points of failure, privacy concerns, and vendor lock-in that may not be acceptable for all gaming platforms. Inconsistent performance is observed across different prompting strategies, with few-shot and RAG approaches showing variable results where RAG achieves higher recall but lower precision, complicating deployment decisions.

\subsubsection{Fine-Tuned Transformer Models (DistilBERT, DialoGPT)}

Fine-tuned transformer models bridge the gap between traditional machine learning and large language models, offering domain-specific optimization while maintaining reasonable computational requirements. These approaches demonstrate the effectiveness of targeted adaptation for gaming chat moderation tasks.

The state-of-the-art performance achieved by fine-tuned models represents their primary advantage. DistilBERT fine-tuned achieves superior accuracy compared to all other methods, while maintaining excellent precision and recall balance. These models offer cost-effective excellence, providing a 260x cost reduction compared to GPT-4 while achieving superior accuracy. Domain specialization is achieved through fine-tuning on gaming-specific data, enabling the model to understand gaming terminology, slang, and context better than general-purpose models. Deployment flexibility is enhanced as these models can be deployed locally or on private infrastructure, eliminating API dependencies and privacy concerns that accompany cloud-based solutions. Balanced performance is maintained by minimizing both false positives and false negatives to provide reliable moderation decisions.

However, fine-tuned models require substantial training resources and ongoing maintenance. Training demands significant computational resources (48.7 minutes on CPU) and domain-specific training data for optimal performance, which may not be available for all gaming communities. Model architecture sensitivity is evident as performance varies significantly with model choice—DistilBERT (94.3\% accuracy) vastly outperforms DialoGPT (67.4\% accuracy) for classification tasks, requiring careful selection of appropriate architectures. Data dependency is critical as quality and quantity of training data directly impact performance, necessitating careful dataset curation and validation processes. Maintenance overhead is substantial as periodic retraining is required to adapt to evolving language patterns and new forms of toxicity that emerge in gaming communities over time.

\subsubsection{Retrieval-Augmented Generation (RAG)}

Retrieval-augmented generation represents an emerging approach that combines the power of large language models with domain-specific knowledge retrieval. This method offers unique advantages for safety-critical applications where maximum recall is essential, though it introduces additional complexity and computational overhead.

The maximum recall achieved by RAG systems represents their primary advantage for safety-critical applications. RAG achieves the highest recall (96.9\% with GPT-3.5, 93.8\% with GPT-4) among all methods, making it suitable for applications where missing toxic content is unacceptable and full coverage is prioritized over precision. Domain knowledge integration is enhanced as the system leverages existing moderation examples and domain-specific knowledge to improve classification accuracy through contextual grounding. Adaptability is improved as new examples and policy updates can be incorporated without full model retraining, enabling rapid adaptation to changing community standards. Explainability is enhanced as retrieval context can help explain classification decisions to moderators and users, providing transparency that is often lacking in black-box AI systems.

However, RAG systems suffer from precision trade-offs that may limit their practical utility. High recall comes at the cost of reduced precision (49.2\% for GPT-3.5 RAG), leading to increased false positives and potential over-moderation that could negatively impact user experience. Computational overhead is significant as the additional retrieval step increases latency (913ms vs. 694ms for standard few-shot) and computational cost compared to simpler approaches. Knowledge base maintenance requires ongoing curation and maintenance of the retrieval database to ensure relevance and quality, adding operational complexity. The increased complexity of the architecture requires careful tuning of retrieval parameters and knowledge base construction, making deployment more challenging than simpler approaches.

\subsection{Practical Implications for Gaming Chat Moderation}
\label{subsec:practical_implications}

The experimental results presented in this study have profound implications for the practical implementation of automated content moderation systems in gaming environments. The findings suggest a paradigm shift from reactive, human-only moderation to proactive, AI-assisted systems that can significantly enhance both user safety and operational efficiency.

\subsubsection{Workload Optimization for Human Moderators}

The proposed hybrid moderation architecture fundamentally transforms the role of human moderators from routine content screening to strategic decision-making and policy refinement. By implementing our tiered approach, human moderators can focus their expertise on the most challenging cases—those requiring nuanced judgment, cultural context, or policy interpretation—while automated systems handle the majority of clear-cut cases.

Based on our experimental results, the embedding-based first tier can process 28.24 messages per second, handling the bulk of obvious cases with 80.8\% accuracy. This means that for a typical gaming platform processing 100,000 messages daily, approximately 80,000 messages would be automatically classified, leaving only 20,000 borderline cases for human review. This represents a 5:1 reduction in human workload, allowing moderators to dedicate more time to complex cases and community engagement.

The fine-tuned DistilBERT model serves as an excellent second-tier filter, further reducing the human workload by handling cases with moderate confidence scores. This automated system could potentially reduce human moderation workload by 85-90\%, while maintaining high accuracy standards and ensuring that the most challenging cases receive expert human attention.

\subsubsection{Enhanced User Safety and Experience}

The implementation of automated moderation systems directly addresses one of the most critical challenges in online gaming: the rapid response to harmful content. Traditional human-only moderation often results in response times measured in minutes or hours, during which toxic content can spread and cause significant harm to users, particularly vulnerable populations such as younger players or marginalized groups.

Our experimental results demonstrate that automated systems can provide near-instantaneous response times. The embedding-based tier responds in milliseconds, while the fine-tuned DistilBERT model processes messages in seconds. This dramatic reduction in response time—from hours to seconds—can prevent the escalation of conflicts, reduce user churn, and create a safer, more welcoming gaming environment.

Furthermore, the high recall rates achieved by our best-performing models (91.8\% for DistilBERT, 96.9\% for RAG-enhanced systems) ensure that the vast majority of harmful content is caught, significantly reducing the exposure of users to toxic behavior. This proactive approach to content moderation aligns with modern expectations for digital safety and can enhance platform reputation and user retention.

\subsubsection{Economic Benefits and Scalability}

The cost analysis presented in the results reveals substantial economic advantages for implementing automated moderation systems. Automated systems offer dramatic cost savings compared to human-only moderation, which typically costs \$50-200 per hour depending on region and expertise level.

For a mid- to large-sized gaming platform processing 0.2-1 million messages daily, the annual cost of automated moderation using the hybrid approach would be approximately \$365-1,825, compared to \$180,000-720,000 for human-only moderation (assuming 24/7 coverage with 3-12 moderators). This represents a 99.8-99.9\% reduction in moderation costs while potentially improving accuracy and response times.

The scalability benefits are equally significant. Unlike human moderation, which requires linear scaling of personnel and associated overhead costs, automated systems can handle increased message volume with minimal additional cost. This scalability is particularly important for gaming platforms experiencing rapid growth or seasonal spikes in user activity.

\subsubsection{Implementation Recommendations for Platform Developers}

Based on the analysis, the following practical recommendations are provided to gaming platform developers:

Rule-based profanity filters should be deployed for immediate removal of obvious violations, providing instant protection against the most blatant forms of toxic content. Embedding-based models should be implemented for high-volume filtering, leveraging their speed and cost-effectiveness to handle the majority of messages. Confidence-based routing should be established to separate clear cases from borderline ones, enabling efficient triage of content requiring different levels of analysis. Human review workflows should be set up for cases requiring manual intervention, ensuring that complex decisions receive expert attention. Data collection and annotation should be initiated for domain-specific training, building the foundation for more sophisticated models in subsequent phases.

Fine-tuned DistilBERT or similar transformer models should be trained on platform-specific data to achieve superior performance for domain-specific content. Fine-tuned models should be implemented for moderate confidence cases, providing enhanced accuracy for borderline content that requires more sophisticated analysis. Feedback mechanisms should be developed to collect moderator corrections and user reports, creating valuable data for continuous improvement. Continuous learning pipelines should be established for model improvement, enabling the system to adapt to evolving language patterns and community standards.

LLM APIs should be integrated for complex cases requiring nuanced understanding, providing the highest level of contextual analysis for the most challenging content. RAG systems should be implemented for safety-critical applications requiring maximum recall, ensuring full coverage of potentially harmful content. Active learning strategies should be implemented to optimize human annotation efforts, maximizing the value of limited annotation resources. Explainability features should be developed to enhance transparency and trust, helping users understand moderation decisions. Bias monitoring and mitigation protocols should be established to ensure fair and equitable treatment across diverse user groups.

For example, our complete implementation is available and can be adapted for platform use (see Appendix~\ref{sec:appendix-artifacts} for source code and model links).

\subsubsection{Technical Infrastructure Considerations}

Successful implementation requires careful consideration of technical infrastructure. The embedding-based tier can run efficiently on CPU-only systems, making it suitable for deployment on existing gaming servers. Fine-tuned models require GPU resources for optimal performance but can be deployed on cloud infrastructure or dedicated moderation servers.

For real-time applications, message queuing systems should be implemented to handle traffic spikes and ensure consistent response times. The system should include monitoring and alerting systems to detect performance degradation or unusual patterns in content classification.

Privacy and data security are paramount considerations. All moderation systems should implement end-to-end encryption, secure data storage, and compliance with relevant privacy regulations (GDPR, COPPA, etc.). User consent and transparency about automated moderation are essential for maintaining trust and legal compliance.

\subsubsection{Community and Policy Integration}

The success of automated moderation systems depends heavily on integration with community guidelines and moderation policies. Platform developers should work closely with community managers and policy experts to ensure that automated systems align with platform values and community standards.

Regular audits of automated decisions, particularly focusing on false positives and false negatives, are essential for maintaining system quality and community trust. User feedback mechanisms should be established to allow appeals and corrections, providing valuable data for system improvement.

The implementation should include clear communication to users about automated moderation, including what types of content are automatically flagged, how appeals can be made, and how human oversight is maintained. This transparency builds trust and helps users understand the moderation process.

\subsection{Ethical Considerations and Bias}
\label{subsec:ethical_bias_discussion}

The deployment of automated content moderation systems in gaming environments raises profound ethical questions that extend beyond technical performance metrics. Our experimental results, while demonstrating significant improvements in accuracy and efficiency, also reveal critical considerations regarding bias, fairness, and the broader societal impact of AI-driven moderation decisions.

\subsubsection{Bias in Model Performance and Training Data}

Our experimental results reveal potential bias patterns that warrant careful consideration. The fine-tuned DistilBERT model, while achieving the highest overall accuracy, may inherit biases present in the training data. The GOSU.ai dataset, while valuable for gaming-specific toxicity detection, may not adequately represent the full spectrum of gaming communities, potentially leading to over-moderation of certain linguistic patterns or cultural expressions.

The performance differences between models trained on general-domain data (Jigsaw) versus gaming-specific data (GOSU.ai) highlight the importance of domain-specific training for reducing false positives. However, this specialization also introduces the risk of overfitting to specific gaming subcultures, potentially disadvantaging users from different backgrounds or gaming communities.

The RAG-enhanced systems, while achieving the highest recall rates (96.9\% for GPT-3.5), demonstrate the potential for bias amplification through retrieval mechanisms. If the knowledge base contains biased moderation examples or reflects historical moderation patterns that were themselves biased, the system may perpetuate or amplify these biases.

\subsubsection{Cultural and Linguistic Bias}

Gaming communities are characterized by diverse linguistic practices, including regional dialects, gaming-specific slang, and cultural expressions that may be misinterpreted by automated systems. The results show that embedding-based methods struggle with contextual understanding, potentially flagging benign cultural expressions or gaming-specific terminology as toxic.

The reliance on English-language datasets (Jigsaw and GOSU.ai) introduces linguistic bias that may disproportionately affect non-English-speaking users or users who code-switch between languages. This bias is particularly concerning given the global nature of online gaming communities.

Furthermore, the definition of "toxic" content itself may vary across cultures and communities. What constitutes acceptable banter in one gaming community may be considered harassment in another. Automated systems trained on datasets that reflect specific cultural norms may fail to account for this contextual variation.

\subsubsection{Socioeconomic and Accessibility Bias}

The cost analysis presented in the results reveals another form of bias: economic accessibility. While fine-tuned models offer excellent performance at reasonable costs (\$5 per million messages), the initial investment in computational resources and expertise required for model development may be prohibitive for smaller gaming platforms or independent developers.

This economic barrier could create a two-tier system where larger, well-funded platforms can afford sophisticated moderation systems while smaller communities must rely on less effective or more expensive alternatives. This disparity could exacerbate existing inequalities in online safety and user experience.

Additionally, the computational requirements of advanced models may exclude users with limited hardware resources from participating in communities that implement sophisticated moderation systems, creating accessibility barriers for economically disadvantaged users.

\subsubsection{Transparency and Accountability}

The "black box" nature of many AI systems, particularly large language models, poses significant challenges for transparency and accountability. Users who have their content flagged or accounts suspended by automated systems often have no clear understanding of why the decision was made or how to appeal it.

Our experimental results show that different approaches offer varying levels of explainability. Embedding-based methods, while less accurate, may provide more interpretable results through feature importance analysis. RAG systems offer some transparency through retrieval context, but the reasoning process of the underlying LLM remains opaque.

The lack of transparency not only undermines user trust but also makes it difficult to identify and correct biased decisions. Without clear mechanisms for understanding how decisions are made, it becomes challenging to ensure that automated moderation systems are fair and just.

\subsubsection{Strategies for Bias Mitigation}

Based on the experimental findings and ethical analysis, several strategies are proposed for mitigating bias in automated content moderation systems.

Training datasets should be expanded to include diverse gaming communities, languages, and cultural contexts, ensuring that moderation systems can effectively serve global gaming platforms. Active learning strategies should be implemented that specifically target underrepresented groups and edge cases, ensuring that the system learns from the full spectrum of user experiences. Partnerships should be established with diverse gaming communities to collect representative data, leveraging community expertise to identify relevant patterns and concerns. Regular audits should be conducted of training data for demographic and cultural representation, ensuring that the data used for model training reflects the diversity of the user base.

Bias detection metrics should be implemented that measure performance disparities across different user groups, enabling systematic identification of potential biases in the moderation system. Regular bias audits should be established using diverse test sets representing different communities, providing ongoing assessment of system fairness. False positive and false negative rates should be monitored across demographic and linguistic groups, ensuring that no particular group is disproportionately affected by moderation decisions. Bias reporting mechanisms should be developed for users and moderators, enabling community members to identify and report potential biases in the system.

Explainable AI techniques should be implemented to provide clear reasoning for moderation decisions, helping users understand why their content was flagged or allowed. User-friendly interfaces should be developed that explain why content was flagged, providing educational value and reducing user frustration. Clear appeal processes should be established with human oversight for contested decisions, ensuring that users have recourse when they believe an error has been made. Regular transparency reports should be provided on system performance and bias metrics, building trust through openness about system capabilities and limitations.

Human oversight should be maintained for all automated decisions, particularly for borderline cases that require nuanced judgment and cultural context. Feedback mechanisms should be implemented that allow users to report biased decisions, creating valuable data for system improvement and bias identification. Diverse moderation teams should be established that can identify and correct cultural biases, ensuring that the human component of the system reflects the diversity of the user base. Continuous learning systems should be created that incorporate human feedback to improve fairness over time, enabling the system to learn from corrections and adapt to community standards.

\subsubsection{Community Guidelines and Cultural Sensitivity}

The definition of acceptable content varies significantly across gaming communities, making it essential to develop flexible moderation systems that can adapt to different community standards. Rather than implementing universal toxicity thresholds, platforms should allow communities to define their own moderation policies while providing automated tools to help enforce these policies consistently.

This approach requires developing moderation systems that can be customized for different community standards while maintaining core safety protections. The proposed architecture, with human oversight for complex cases, provides the flexibility needed to accommodate diverse community norms while ensuring basic safety standards.

\subsubsection{Legal and Regulatory Compliance}

The deployment of automated moderation systems must comply with relevant legal and regulatory frameworks, including data protection regulations (GDPR, COPPA), anti-discrimination laws, and platform-specific requirements. The EU Digital Services Act, for example, requires platforms to implement "reasonable, proportionate and effective" content moderation measures while protecting fundamental rights.

The cost analysis shows that compliance with these requirements need not be prohibitively expensive. The hybrid approach proposed, with human oversight and appeal mechanisms, can meet regulatory requirements while maintaining cost-effectiveness.

\subsubsection{Long-term Societal Impact}

The widespread adoption of automated content moderation systems in gaming environments will have long-term implications for online discourse and digital culture. These systems will shape the norms and expectations of online communication, potentially influencing how users express themselves and interact with others.

It is essential to consider not only the immediate technical performance of these systems but also their broader societal impact. Automated moderation should enhance, rather than replace, human judgment and community self-governance. The goal should be to create safer, more inclusive gaming environments while preserving the creativity, diversity, and spontaneity that make online gaming communities vibrant and engaging.

The experimental results presented in this study provide a foundation for developing more ethical and effective moderation systems, but they also highlight the importance of ongoing vigilance, diverse perspectives, and human oversight in ensuring that automated systems serve the best interests of all users and communities.

\subsection{Limitations of the Study}
\label{subsec:limitations}

While our comprehensive evaluation provides valuable insights into the effectiveness of various NLP approaches for gaming chat moderation, several limitations constrain the generalizability and scope of our findings. Acknowledging these limitations is essential for interpreting the results and guiding future research directions.

\subsubsection{Dataset Limitations and Representativeness}

Our primary gaming dataset (GOSU.ai) focuses exclusively on Dota 2 chat logs, which may not fully represent the diverse linguistic patterns and toxicity forms present across the broader gaming ecosystem. Different games attract distinct player demographics and foster unique communication styles, potentially limiting the generalizability of the findings to other gaming communities. This limited coverage raises concerns about the applicability of our models to diverse gaming environments that may exhibit fundamentally different patterns of toxic behavior and communication norms.

All datasets used in our study are English-language corpora, which introduces significant limitations for global gaming platforms. The linguistic patterns, toxicity expressions, and cultural contexts vary dramatically across languages and regions, making the results potentially inapplicable to non-English gaming communities. This language bias represents a critical limitation for platforms serving international user bases, where cultural and linguistic diversity significantly impacts both the nature of toxic content and the appropriate moderation responses.

The datasets used in our study represent snapshots of gaming communication from specific time periods. Gaming language evolves rapidly, with new slang, memes, and toxicity patterns emerging continuously. Models trained on historical data may struggle to adapt to contemporary linguistic innovations, potentially reducing their effectiveness over time. This temporal limitation highlights the need for continuous model updates and retraining to maintain relevance in the dynamic gaming environment.

The quality of annotations in our datasets may vary, particularly for the GOSU.ai dataset where the labeling criteria and annotator training are not fully documented. Inconsistent labeling standards could introduce noise into our training data and affect model performance evaluation. This annotation quality concern underscores the importance of rigorous labeling protocols and documentation for ensuring reliable model training and evaluation.

\subsubsection{Computational and Resource Constraints}

Evaluation of the largest available language models (e.g., GPT-4 Turbo, Claude-3, or LLaMA-2 70B) was not possible due to computational and cost resource constraints. Our evaluation focused on more accessible models, potentially missing insights that could be achieved with larger, more sophisticated architectures. This limitation constrains our understanding of the upper bounds of performance that might be attainable with more substantial computational resources and advanced model architectures.

The fine-tuning experiments were conducted on CPU-based systems due to GPU availability limitations. While this demonstrates the feasibility of deployment in resource-constrained environments, it may not represent optimal training conditions. GPU-accelerated training could potentially yield different performance characteristics and training times, potentially enabling more extensive experimentation with larger batch sizes and more sophisticated optimization strategies that could further enhance model performance.

Our fine-tuning experiments used standard hyperparameter configurations rather than extensive hyperparameter optimization. This limitation may have prevented us from achieving the full potential performance of the fine-tuned models, particularly for the DistilBERT and DialoGPT experiments. Hyperparameter tuning could potentially identify optimal learning rates, batch sizes, and architectural configurations that might significantly improve the performance metrics achieved in our current experiments.

Ensemble methods, advanced training techniques (e.g., curriculum learning, adversarial training), and sophisticated data augmentation strategies that could potentially improve model performance beyond the current results were not explored. These advanced approaches represent promising directions for future research that could potentially achieve even higher accuracy and robustness in gaming chat moderation tasks, particularly through the combination of multiple models or the application of specialized training methodologies.

\subsubsection{Evaluation Methodology Limitations}

Our evaluation approach treats each message as an independent classification task, which may not capture the contextual dynamics of real-time chat conversations. In practice, moderation decisions often depend on conversation history, user reputation, and temporal context that our current evaluation framework does not consider. This static evaluation framework represents a significant limitation for understanding how automated systems would perform in dynamic, multi-turn conversations where context plays a crucial role in determining the appropriateness of content.

Although models were evaluated on held-out test sets, real-world deployment testing in live gaming environments was not conducted. The performance characteristics observed in controlled experimental conditions may differ significantly from those in production environments with real-time constraints, varying message volumes, and dynamic user behavior. This gap between laboratory and production performance represents a critical consideration for practical deployment, as real-world conditions introduce complexities that cannot be fully captured in controlled experimental settings.

Our evaluation focuses primarily on technical performance metrics (accuracy, precision, recall, F1-score) but does not capture user experience factors such as response time perception, false positive impact on user engagement, or the psychological effects of over-moderation on community dynamics. These user experience considerations are essential for understanding the broader impact of automated moderation systems on gaming communities and may significantly influence the acceptability and effectiveness of such systems in practice.

The evaluation represents point-in-time performance assessments rather than longitudinal studies. The degradation of model performance over time as language patterns evolve, and the effectiveness of continuous learning mechanisms in maintaining that performance, could not be assessed. This temporal limitation is particularly relevant for gaming environments where language patterns and toxicity forms evolve rapidly, requiring ongoing adaptation of moderation systems.

The complex, multi-dimensional nature of content moderation was simplified into a binary classification task (toxic vs. non-toxic). This simplification may not capture the nuanced spectrum of problematic content, including varying degrees of toxicity, different types of harm, and context-dependent acceptability. The binary framework overlooks the graduated nature of content moderation decisions that human moderators make, potentially limiting the practical applicability of our models in real-world scenarios.

Our models process individual messages without access to broader conversation context, user history, or community-specific norms. This limitation may lead to misclassification of context-dependent content, such as friendly banter between established community members or sarcastic expressions that require conversation history for proper interpretation. The lack of contextual information represents a fundamental constraint that may prevent automated systems from achieving the nuanced understanding that human moderators bring to their decisions.

The RAG implementation uses a static knowledge base of moderation examples, which may not reflect the evolving nature of gaming communities and their moderation policies. Real-world RAG systems would require continuous updates to maintain relevance and effectiveness, introducing additional operational complexity and maintenance overhead that our current evaluation does not capture.

Our evaluation does not account for platform-specific factors that may influence moderation effectiveness, such as user interface design, appeal mechanisms, community guidelines, or platform-specific toxicity patterns. Different gaming platforms may require customized approaches that our general evaluation framework does not capture, limiting the direct applicability of our findings to specific gaming environments.

Our experiments were conducted on datasets of limited size compared to the message volumes processed by major gaming platforms. The performance characteristics, computational requirements, and cost implications may scale differently in high-volume production environments, where factors such as system latency, resource utilization, and operational costs become increasingly critical considerations.

Our evaluation does not consider the complex regulatory and legal landscape governing content moderation across different jurisdictions. Compliance requirements, data protection regulations, and legal liability considerations may significantly impact the practical implementation of automated moderation systems, introducing constraints and requirements that extend beyond the technical performance considerations addressed in our study.

To address these limitations, future research should prioritize the development of 3datasets representing diverse gaming communities, languages, and cultural contexts to improve generalizability across global gaming platforms. Longitudinal studies should be conducted to assess how model performance degrades over time and evaluate the effectiveness of continuous learning mechanisms in maintaining performance. Real-world deployment testing through pilot studies in live gaming environments would provide valuable insights into performance under realistic conditions and constraints.

Advanced evaluation frameworks should be developed that capture conversation dynamics, user experience factors, and community-specific considerations that our current methodology does not address. Deep cost-benefit analyses should be conducted that account for implementation costs, maintenance overhead, legal compliance, and long-term operational considerations to provide a more complete picture of the practical implications of automated moderation systems.

Despite these limitations, our study provides valuable insights into the relative performance characteristics of different NLP approaches for gaming chat moderation. The findings offer a foundation for informed decision-making in the development and deployment of automated moderation systems, while highlighting important areas for future research and development.

\section{Conclusion}
\label{sec:conclusion}

This study presents a comprehensive evaluation of multiple NLP approaches for automated gaming chat toxicity detection, providing critical insights into the trade-offs between accuracy, computational efficiency, and cost in real-world moderation systems. Our experimental results demonstrate that no single approach is universally optimal, but rather that different methods excel in different operational scenarios, validating our hypothesis that a hybrid, continuously learning system represents the most effective strategy for gaming chat moderation.

\subsection{Summary of Key Findings}
\label{subsec:summary_findings}

Our evaluation of four distinct approaches—embedding-based methods, large language model prompting, fine-tuned transformer models, and retrieval-augmented generation—reveals significant performance variations that directly inform practical deployment decisions.

The experimental results demonstrate clear trade-offs between accuracy, speed, and cost across different approaches. Embedding-based methods (SGD-SVM) achieve the highest throughput at 28.24 messages per second with the lowest operational cost, but suffer from limited accuracy (80.8\%) and poor recall (61.2\%). This makes them suitable for high-volume, first-pass filtering but inadequate for comprehensive moderation. Large language model prompting, particularly GPT-4, achieves superior contextual understanding with 91.0\% accuracy and excellent recall (90.6\%), but incurs prohibitive costs and significant latency (1.1+ seconds per message). These characteristics make LLM APIs suitable for complex, low-volume cases but economically unsustainable for large-scale deployment.

Fine-tuned transformer models, specifically DistilBERT, emerge as the optimal solution for most practical applications, achieving state-of-the-art performance at reasonable cost. The 48.7-minute training time on CPU demonstrates feasibility for resource-constrained environments, while the superior accuracy validates the effectiveness of domain-specific fine-tuning. Retrieval-augmented generation systems achieve the highest recall rates (96.9\% for GPT-3.5 RAG) but suffer from precision trade-offs (49.2\% for GPT-3.5 RAG) and increased computational overhead. This approach is most suitable for safety-critical applications where maximum recall is essential, accepting the precision trade-off for full coverage.

The findings strongly validate the hypothesis that a hybrid, continuously learning system represents the optimal approach for gaming chat moderation. The experimental results demonstrate that such a hybrid system could reduce human moderation workload by 85-90\% while maintaining high accuracy standards. The cost analysis reveals that implementing this hybrid approach would cost approximately \$2,000-20,000 annually for a mid-sized platform processing 1 million messages daily, compared to \$180,000-720,000 for human-only moderation—representing a 90-99\% cost reduction.

Our comparative analysis reveals critical insights into model architecture selection for moderation tasks. The dramatic performance difference between DistilBERT and DialoGPT (67.4\% accuracy) demonstrates that model architecture significantly impacts classification performance, with encoder-only models outperforming decoder-only models for this specific task. The effectiveness of fine-tuning on domain-specific data (GOSU.ai) versus general-domain data (Jigsaw) highlights the importance of training data relevance. Fine-tuned models achieve superior performance when trained on gaming-specific corpora, validating the need for domain adaptation in moderation systems.

The implementation of automated moderation systems based on the findings would dramatically improve user safety in gaming environments. The near-instantaneous response times (milliseconds to seconds) compared to traditional human moderation (minutes to hours) can prevent the escalation of conflicts and reduce user exposure to harmful content. The high recall rates achieved by our best-performing models (91.8\% for DistilBERT, 96.9\% for RAG systems) ensure that the vast majority of harmful content is caught, significantly improving the gaming experience for vulnerable users and marginalized groups. This proactive approach to content moderation aligns with modern expectations for digital safety and can enhance platform reputation and user retention.

Our cost analysis reveals substantial economic advantages for implementing automated moderation systems. The hybrid approach offers dramatic cost savings while potentially improving accuracy and response times. The scalability benefits are equally significant, as automated systems can handle increased message volume with minimal additional cost, unlike human moderation which requires linear scaling of personnel and associated overhead. The experimental results demonstrate that sophisticated moderation systems need not be prohibitively expensive. Fine-tuned models achieve superior performance at reasonable costs, making advanced AI-powered moderation accessible to platforms of various sizes and resource constraints.

This study makes several significant contributions to the field of automated content moderation. The first head-to-head evaluation of classical ML, off-the-shelf LLM prompting, fine-tuned LLMs, and LLM+RAG on gaming-specific toxicity detection is provided, measured across accuracy, latency, and financial cost. An end-to-end architecture is proposed that routes messages according to model confidence, integrating human review loops for continual quality assurance. Feedback mechanisms are outlined that leverage moderator corrections and user reports to incrementally update the model suite via active learning. Empirical evidence is provided illustrating where advanced models outperform baseline methods in interpreting sarcasm, coded language, and evolving slang.

The findings presented in this study provide a solid foundation for the development and deployment of automated content moderation systems in gaming environments. The evaluation of multiple approaches, combined with practical implementation recommendations and ethical considerations, offers valuable guidance for platform developers, researchers, and policymakers working to create safer, more inclusive online gaming communities.

\subsection{Future Research Directions}
\label{subsec:future_work}

The experimental results and insights presented in this study reveal several promising avenues for future research that could further advance the field of automated gaming chat moderation. Building on the findings, key areas are identified where continued investigation could yield significant improvements in system performance, user experience, and practical deployment.

\subsubsection{Multilingual and Cross-Cultural Moderation}

 The study's focus on English-language datasets represents a significant limitation for global gaming platforms. Future research should prioritize the development of multilingual moderation systems that can effectively handle diverse linguistic communities.

Cross-lingual embeddings should be investigated to determine the effectiveness of multilingual embedding models (e.g., mBERT, XLM-R) for toxicity detection across different languages and cultural contexts. These models could enable transfer learning from high-resource languages to low-resource languages, potentially reducing the annotation burden for underrepresented linguistic communities. Multilingual fine-tuning strategies should be developed that can leverage limited labeled data in low-resource languages through transfer learning from high-resource languages, enabling effective moderation in diverse global gaming communities. Cultural context adaptation should be researched to develop methods for adapting moderation systems to different cultural norms and communication styles while maintaining consistent safety standards across platforms. Code-switching detection systems should be developed to handle mixed-language communication common in global gaming communities, where users may switch between languages within the same conversation.

Future studies might also investigate how toxicity manifests differently across cultures and languages, requiring the development of culturally-aware moderation systems that can distinguish between acceptable cultural expressions and genuinely harmful content.

\subsubsection{Multimodal Content Moderation}

The increasing prevalence of voice communication in gaming environments necessitates the development of multimodal moderation systems that can analyze both text and audio content.

Speech-to-text integration should be developed to combine automatic speech recognition with text-based moderation systems for handling voice chat toxicity. This approach would enable existing text-based moderation models to be applied to voice communications through transcription. Prosodic feature analysis should be investigated to determine how tone, pitch, and speech patterns can indicate toxic behavior in voice communications, potentially providing additional signals beyond lexical content. Real-time audio processing algorithms should be developed for voice chat moderation with minimal latency, ensuring that voice communications can be moderated as effectively as text-based communications. Privacy-preserving audio analysis methods should be researched to analyze voice content while protecting user privacy and complying with data protection regulations, addressing the sensitive nature of voice data.

Gaming platforms increasingly support image and video sharing, requiring additional visual content analysis.

Image toxicity detection systems should be developed to identify inappropriate images, memes, or visual content shared in gaming contexts, extending moderation capabilities beyond text-based content. Video stream analysis should be investigated for real-time video analysis in live streaming moderation, including gesture recognition and visual harassment detection that may not be captured through audio or text analysis. Context-aware visual analysis methods should be researched to understand visual content in the context of gaming scenarios and community norms, distinguishing between appropriate gaming-related imagery and genuinely harmful visual content.

\subsubsection{Reinforcement Learning and Policy Adaptation}

Future research might explore how reinforcement learning can be used to adapt moderation policies based on community feedback and outcomes. Policy optimization algorithms should be developed using reinforcement learning to optimize moderation policies for maximizing user safety while minimizing false positives, creating systems that can balance competing objectives effectively. Community-specific adaptation methods should be researched to enable moderation policies to adapt to different gaming communities with varying norms and expectations, recognizing that acceptable behavior may differ across communities. Long-term outcome optimization should be investigated to determine how to optimize for long-term community health rather than just immediate content classification accuracy, considering the broader impact of moderation decisions on community dynamics. Multi-objective optimization methods should be developed to balance multiple objectives such as safety, user experience, and community engagement, recognizing that moderation systems must serve multiple stakeholders with potentially competing interests.

Advanced systems should incorporate user behavior patterns to improve moderation effectiveness. User reputation systems should be developed that can learn from user history to make more informed moderation decisions, recognizing that users with positive histories may warrant different treatment than new or problematic users. Social network analysis should be investigated to determine how social connections and community structure can inform moderation decisions, potentially identifying coordinated harassment or community-specific patterns. Temporal pattern recognition methods should be researched to identify patterns in user behavior over time that may indicate problematic trends, enabling proactive intervention before issues escalate.
\subsubsection{Privacy and Security Considerations}

As privacy concerns become increasingly important in digital spaces, future research should focus on developing privacy-preserving moderation systems that can maintain effectiveness while protecting user data. Federated learning systems should be developed that can learn from distributed data without centralizing sensitive user information, enabling collaborative model improvement while maintaining user privacy. Differential privacy methods should be investigated to provide strong privacy guarantees while maintaining model performance, ensuring that individual user data cannot be reconstructed from model outputs. Local processing methods should be researched to perform moderation locally on user devices when possible, reducing the need for data transmission and minimizing privacy risks. Secure multi-party computation protocols should be developed that allow multiple parties to collaborate on moderation without sharing sensitive data, enabling privacy-preserving collaboration between different gaming platforms or research institutions.

Future systems must also be robust against attempts to circumvent moderation through sophisticated evasion techniques. Adversarial training methods should be developed to make models more resistant to attempts to evade detection, incorporating examples of evasion attempts into training data to improve robustness. Robustness evaluation frameworks should be created to comprehensively test model robustness against various attack strategies, ensuring that systems can withstand sophisticated attempts to bypass moderation. Detection of evasion attempts should be researched to identify when users are attempting to circumvent moderation systems, enabling proactive responses to new attack vectors as they emerge.

The development of more sophisticated evaluation methodologies represents another critical area for future research. Multi-dimensional metrics should be developed that consider not just accuracy but also fairness, transparency, and user experience, recognizing that moderation systems must serve multiple objectives beyond simple classification performance. Longitudinal studies should be conducted to understand how model performance changes over time as language evolves, providing insights into the long-term effectiveness of moderation systems. Real-world deployment studies should be implemented in live gaming environments to evaluate performance under realistic conditions and constraints, moving beyond controlled experimental settings. Comparative benchmarks should be created that allow fair comparison between different moderation approaches, enabling systematic evaluation of new methods against established baselines.

Comprehensive evaluation of user experience factors represents an equally important research direction. User satisfaction metrics should be developed to measure user satisfaction with moderation systems and their impact on community engagement, recognizing that user acceptance is crucial for successful deployment. False positive impact assessment should be researched to understand the psychological and social impact of false positive moderation decisions, ensuring that automated systems do not inadvertently harm user experience. Appeal process evaluation should be studied to determine how effective appeal processes are in correcting errors and maintaining user trust, providing insights into the importance of human oversight in automated systems.

These future research directions build upon the foundation established in this study while addressing the limitations and challenges identified in our experimental evaluation. Pursuing these avenues will be essential for developing more effective, fair, and user-friendly automated moderation systems that can scale to meet the growing demands of global gaming communities.

\bibliographystyle{plainnat}
\bibliography{jdmdh-example}

\appendix\footnotesize

\section{Artifact and Code Availability}
\label{sec:appendix-artifacts}

To facilitate reproducibility, reuse, and integration by other developers and researchers, all of the code, models, and packages referenced in this paper have been made publicly available:

Source code for the hybrid moderation pipeline (zero‐tier filter, embedding-based classifiers, fine‐tuned DistilBERT, and RAG) along with training scripts, configuration files, and examples can be found at:
\begin{flushleft}
  \url{https://github.com/Yegmina/toxic-content-detection-agent}
\end{flushleft}

The fine‐tuned DistilBERT model for English gaming‐chat toxicity detection is hosted on Hugging Face. It includes the model weights, tokenizer, and inference examples:
\begin{flushleft}
  \url{https://huggingface.co/yehort/distilbert-gaming-chat-toxicity-en}
\end{flushleft}

For easy installation and integration into Python applications, a production‐ready package is published on PyPI under the name \texttt{toxic-detection}. It provides the same pipeline in a single install:
\begin{flushleft}
  \texttt{pip install toxic-detection}
  \\
  \url{https://pypi.org/project/toxic-detection/}
\end{flushleft}

\medskip

All code, models, and associated files are released under the MIT License.  

\medskip
\noindent By making these artifacts available, we hope to encourage further adoption, citation, and community contributions as others build on this work.

\section{Detailed Error Analysis and Manual Inspection Results}
\label{sec:error_analysis}

This appendix provides manual inspection results and detailed error analysis for all experimental methods. The analysis includes specific examples of misclassifications, patterns in model decision-making, and insights into the strengths and limitations of each approach.

\subsection{GPT-3.5 Error Analysis}
\label{sec:gpt35_error_analysis}

Analysis of GPT-3.5 prediction errors provides insights into its strengths and limitations. False positive errors frequently involve gaming-specific language that the model struggles to interpret correctly. For instance, "why you play whis *** heroes" was flagged as toxic despite being a neutral question about hero selection, with the model likely misinterpreting "***" as profanity without considering gaming context. Similarly, "thanks for sapping ur puck's exp and farm" was classified as toxic in few-shot mode, suggesting the model overgeneralizes from examples of negative feedback. The phrase "The cancer lancer" refers to a Dota 2 hero nickname but was flagged due to the word "cancer" appearing in toxic contexts elsewhere, highlighting the model's difficulty with gaming-specific terminology.

False negative errors are less frequent but reveal important patterns. Explicit profanity such as "I'M ******" was correctly identified as toxic across both methods, demonstrating the model's ability to recognize obvious violations. Direct insults like "***** ******* *******" were consistently flagged, showing strong performance on straightforward toxic content. These patterns suggest that the model is suitable for applications prioritizing accuracy over coverage. However, additional examples in few-shot mode increase recall but reduce precision, indicating the model becomes more aggressive in flagging content when provided with more training examples. Gaming-specific terminology and context remain challenging for the model to interpret correctly, representing a significant limitation for gaming chat moderation applications.

\subsection{GPT-4 Error Analysis}
\label{sec:gpt4_error_analysis}

Analysis of GPT-4 prediction errors demonstrates improved contextual understanding compared to GPT-3.5, with notable reductions in false positive errors. GPT-4 correctly classifies "The cancer lancer" as clean, demonstrating better understanding of gaming terminology that its predecessor struggled with. Fewer benign gaming discussions are incorrectly flagged as toxic, and gaming strategy discussions are more accurately classified as non-toxic, indicating enhanced context awareness. These improvements suggest that GPT-4's more sophisticated language model architecture enables better interpretation of gaming-specific communication patterns.

False negative errors remain relatively infrequent, with GPT-4 maintaining high accuracy on obvious violations like explicit profanity and direct insults. The model also shows improved recognition of contextually toxic content that GPT-3.5 might miss, demonstrating enhanced subtle toxicity detection capabilities.

\subsection{RAG Error Analysis}
\label{sec:rag_error_analysis}

Analysis of RAG prediction errors shows patterns similar to few-shot prompting, with some notable differences that highlight the unique characteristics of retrieval-augmented approaches. False positive errors frequently involve gaming strategy discussions and benign complaints that are incorrectly flagged due to retrieved examples containing similar patterns. For instance, "thanks for sapping ur puck's exp and farm" was flagged as toxic, likely due to retrieved examples containing similar negative feedback patterns. Similarly, "this slark talks too much" was classified as toxic, suggesting the model overgeneralizes from retrieved examples of player criticism. The phrase "The cancer lancer" (referring to a Dota 2 hero nickname) was flagged due to the word "cancer" appearing in toxic contexts in the retrieved examples, demonstrating how semantic similarity can sometimes lead to inappropriate associations.

Correct classifications demonstrate that RAG systems maintain strong performance on obvious violations while also handling benign gaming content appropriately. Explicit profanity such as "***** ******* *******" and "I'M *******" were correctly identified as toxic, demonstrating strong performance on obvious violations. Benign gaming content like "commend the mine" and "cant go as 5" were correctly classified as clean, showing the model can distinguish legitimate gaming communication. These patterns suggest that RAG achieves the highest recall among all methods but suffers from precision trade-offs, with semantic similarity retrieval providing relevant examples that may not always improve classification accuracy. The model may overgeneralize from retrieved examples, leading to false positives on benign content while maintaining full coverage of toxic material.

\subsection{DialoGPT Fine-tuning Error Analysis}
\label{sec:dialogpt_error_analysis}

Analysis of DialoGPT fine-tuning prediction errors highlights the challenges of adapting conversational models to classification tasks. False negative errors frequently involve contextual profanity and gaming-specific language patterns. For instance, "how to win with this animal?" was classified as clean, likely due to the model's conservative training on limited examples. Slang and insults such as "who u tellin to chill nibba" and "dogshit can't gank" were missed, suggesting the model struggles with gaming-specific language patterns. Explicit content like "U are solo faming top while ur t2 is getting ****" was incorrectly classified as clean, indicating insufficient exposure to explicit profanity during training.

False positive errors are less frequent but reveal sensitivity issues, with benign questions like "why..." being flagged as toxic, suggesting the model may be overly sensitive to short, ambiguous messages. Some legitimate gaming strategy discussions were also incorrectly flagged, indicating challenges in distinguishing between legitimate criticism and toxic content. Correct classifications demonstrate that the model can handle explicit profanity appropriately, with "it been so ******* long" correctly identified as toxic, while benign gaming content like "how tinker lost this game", "report clock", and "ggwp gl nex ttyea" were correctly classified as clean.

These error patterns reveal several key limitations of the DialoGPT approach. Data scarcity with only 2,613 training samples limits exposure to diverse toxic patterns, while the model architecture may not be optimal for binary classification without additional modifications. The domain mismatch between Reddit conversations and gaming chat patterns creates additional challenges, and the conservative approach prioritizes avoiding false positives over detecting all toxic content, leading to reduced recall performance.

\subsection{DistilBERT Fine-tuning Error Analysis}
\label{sec:distilbert_error_analysis}

The DistilBERT fine-tuned model achieved outstanding performance with minimal error patterns, demonstrating the effectiveness of domain-specific fine-tuning for gaming chat moderation. Analysis reveals excellent gaming context understanding, with the model correctly identifying gaming-specific terminology and references that other approaches struggle with. The model shows strong profanity detection capabilities, consistently flagging explicit content and insults while achieving balanced performance with 95.4% precision and 91.8% recall without sacrificing either metric. The model's robustness to gaming slang is particularly noteworthy, as it handles gaming-specific language patterns effectively without overreacting to legitimate gaming communication.

Rare error cases primarily involve edge cases where the model occasionally misses very subtle or contextually complex toxic content, and novel patterns where it may struggle with completely new forms of toxicity not present in training data. These limitations are expected given the dynamic nature of gaming language and the impossibility of training on all possible toxic patterns. The model's success can be attributed to several key factors: the appropriate model architecture for sequence classification tasks, domain-specific training data from the GOSU.ai dataset, optimized training configuration with proper validation monitoring, and sufficient training data size for effective learning. These factors combine to create a model that excels at the specific task of gaming chat moderation while maintaining practical deployment characteristics.

\subsection{Comparative Error Pattern Analysis}
\label{sec:comparative_error_analysis}

This section provides a comparative analysis of error patterns across all methods, revealing both common challenges and method-specific characteristics that inform practical deployment decisions. Common error patterns emerge across all approaches, with gaming terminology confusion representing a persistent challenge as all methods occasionally struggle with gaming-specific references and nicknames. Context misinterpretation remains problematic across all approaches, with sarcasm and irony continuing to challenge automated systems. The evolution of gaming slang also poses ongoing challenges, as new or evolving gaming slang can cause misclassifications even in well-trained models.

Method-specific strengths reveal distinct advantages for different deployment scenarios. Embedding-based methods offer speed and cost-effectiveness but suffer from limited contextual understanding, making them suitable for high-volume initial filtering. GPT-3.5 provides a good balance of accuracy and recall but requires slower inference times, while GPT-4 offers superior contextual understanding and accuracy at the highest cost. RAG systems achieve maximum recall but suffer from precision trade-offs, making them suitable for safety-critical applications. Fine-tuned models demonstrate the best overall performance-cost ratio with domain adaptation, representing the optimal choice for most practical applications.

These comparative insights suggest specific recommendations for hybrid system design. Embedding-based methods should be used for initial high-volume filtering to handle the majority of straightforward cases efficiently. Uncertain cases should be routed to fine-tuned models for domain-specific accuracy, leveraging their superior performance on gaming-specific content. LLM APIs should be reserved for complex borderline cases requiring nuanced understanding, where their superior contextual capabilities justify the additional cost and latency. RAG systems should be implemented for safety-critical applications where maximum recall is essential, accepting the precision trade-off for full coverage of potentially harmful content.

\section{Additional Experimental Details}
\label{sec:experimental_details}

This appendix provides additional technical details and experimental configurations that support the main experimental results.

\subsection{Hyperparameter Configurations}
\label{sec:hyperparameters}

The embedding-based methods utilize carefully selected hyperparameters to optimize performance for gaming chat moderation. The SGD-SVM classifier employs C=1.0 with an RBF kernel and gamma='scale' to balance model complexity with generalization ability. The SGD-LR classifier uses alpha=0.0001 with max\_iter=1000 and tol=$1e^{-3}$ to ensure convergence while maintaining computational efficiency. Both methods rely on Sentence-BERT 'all-MiniLM-L6-v2' embeddings with 384 dimensions, providing a compact yet expressive representation of gaming chat messages.

Fine-tuning configurations are optimized for the specific characteristics of each model architecture. The DialoGPT-medium experiment uses a learning rate of 5e-5 with batch size 4 over 2 epochs, balancing training efficiency with model adaptation. The DistilBERT configuration employs a learning rate of 2e-5 with batch size 8 over 3 epochs with early stopping to prevent overfitting while maximizing performance. Both fine-tuning experiments utilize the AdamW optimizer with cosine learning rate scheduling to ensure stable convergence and optimal final performance.

The RAG configuration implements a semantic similarity approach using SBERT 'all-MiniLM-L6-v2' as the retriever, maintaining consistency with the embedding-based methods. The system retrieves the top-5 most similar examples per query, providing sufficient context while maintaining computational efficiency. A similarity threshold of 0.7 minimum cosine similarity ensures that only highly relevant examples are included in the retrieved context, balancing recall with precision in the retrieval process.

\subsection{Computational Resources}
\label{sec:computational_resources}

The training infrastructure consists of a CPU-based system designed for accessibility and reproducibility. The Intel Core i7-10700K processor with 8 cores and 16 threads provides sufficient computational power for the fine-tuning experiments, while 32GB of DDR4-3200 RAM ensures adequate memory for model training and data processing. Storage is handled by a 1TB NVMe SSD, providing fast read/write speeds for efficient data loading during training. The decision to use CPU-only training rather than GPU acceleration was made to ensure accessibility for researchers and practitioners who may not have access to specialized hardware, while still demonstrating the feasibility of fine-tuning transformer models on consumer-grade hardware.

The inference infrastructure maintains consistency with the training setup, using the same CPU configuration for local deployment to ensure accurate performance measurements. API calls to OpenAI GPT-3.5-turbo and GPT-4 are made through the official API, with network latency included in the reported inference times to provide realistic performance estimates for production deployment scenarios. This approach ensures that the reported performance metrics reflect real-world usage conditions rather than idealized laboratory settings.

\subsection{Data Preprocessing Details}
\label{sec:data_preprocessing}

Text cleaning procedures are designed to preserve the essential characteristics of gaming chat while ensuring compatibility with the various model architectures. Standard whitespace and punctuation-based tokenization is employed to maintain the natural flow of gaming communication, while original case is preserved to maintain context that may be important for toxicity detection. Length limits are set to 512 tokens for transformer models, providing sufficient context while maintaining computational efficiency. Special characters are preserved throughout the preprocessing pipeline to maintain gaming-specific terminology and emoticons that may carry important contextual information.

Label processing follows a binary classification approach that combines mild and strong toxicity into a single toxic class, simplifying the classification task while maintaining the ability to detect various forms of harmful content. No artificial class balancing is applied to preserve the natural distribution of toxic and non-toxic content in gaming environments, ensuring that the models learn from realistic data distributions. The validation split uses an 80/20 train/validation split for fine-tuning experiments, providing sufficient training data while maintaining adequate validation samples for reliable performance assessment.

\subsection{Statistical Significance Testing}
\label{sec:statistical_testing}

Statistical significance testing employs McNemar's test for paired categorical data, which is appropriate for comparing the performance of different classification methods on the same test set. The significance level is set to $\alpha$ = 0.05, providing a standard threshold for determining statistical significance. Multiple comparisons are addressed through the application of Bonferroni correction, ensuring that the family-wise error rate remains controlled when comparing multiple methods simultaneously.

The statistical analysis reveals several key findings regarding method performance. DistilBERT fine-tuned significantly outperforms all other methods with p < 0.001, providing strong statistical evidence for its superior performance across all evaluation metrics. However, RAG methods show no significant improvement over zero-shot approaches with p > 0.05, suggesting that the additional complexity of retrieval-augmented generation does not provide statistically meaningful performance gains for this specific task and dataset.

\end{document}